\documentclass[journal]{IEEEtran}
\usepackage{amsmath,amsfonts}
\usepackage{array}
\usepackage{algorithmic}
\usepackage[ruled,lined,linesnumbered,ruled]{algorithm2e}  
\usepackage{multirow}
\usepackage{graphicx} 
\usepackage{float} 
\usepackage[caption=false,font=footnotesize,labelfont=rm,textfont=rm]{subfig}
\usepackage{CJKutf8}
\usepackage{textcomp}
\usepackage{stfloats}
\usepackage{url}
\usepackage{verbatim}
\usepackage{graphicx}
\usepackage{cite}
\usepackage{lipsum}
\begin{document}
\begin{CJK}{UTF8}{gbsn}
\title{\textbf{Recurrent Stochastic Configuration Networks with Hybrid Regularization for Nonlinear Dynamics Modelling}}

\author{
  Gang Dang \\
  State Key Laboratory of Synthetical Automation for Process Industries \\
  Northeastern University, Shenyang 110819, China\\
   Dianhui Wang 
  \thanks{\textit{\underline{Corresponding author}}: 
\textbf{dh.wang@deepscn.com}}\\
  State Key Laboratory of Synthetical Automation for Process Industries \\
  Northeastern University, Shenyang 110819, China \\
 Research Center for Stochastic Configuration Machines\\
  China University of Mining and Technology, Xuzhou 221116, China\\
}
\maketitle
\newtheorem{remark}{\bf Remark}
        
\begin{abstract}                          
Recurrent stochastic configuration networks (RSCNs) have shown great potential in modelling nonlinear dynamic systems with uncertainties. This paper presents an RSCN with hybrid regularization to enhance both the learning capacity and generalization performance of the network. Given a set of temporal data, the well-known least absolute shrinkage and selection operator (LASSO) is employed to identify the significant order variables. Subsequently, an improved RSCN with L2 regularization is introduced to approximate the residuals between the output of the target plant and the LASSO model. The output weights are updated in real-time through a projection algorithm, facilitating a rapid response to dynamic changes within the system. A theoretical analysis of the universal approximation property is provided, contributing to the understanding of the network's effectiveness in representing various complex nonlinear functions. Experimental results from a nonlinear system identification problem and two industrial predictive tasks demonstrate that the proposed method outperforms other models across all testing datasets. 
\end{abstract}
\section{Introduction}
\IEEEPARstart{N}{owdays}, neural networks (NNs) have demonstrated promise in capturing complex patterns and relationships within the process data, making them powerful for modelling nonlinear dynamic systems \cite{ref1,ref2,ref3}. However, in many real-world applications, the order and distribution of online-collected samples remain unknown and constantly varying. These uncertainties significantly impact the model's learning capacity and may lead to a well-trained model underperforming on test samples, resulting in ill-posed problems \cite{ref4,ref5,ref6}. Recurrent neural networks (RNNs) utilize a reservoir to store and update historical state information. This unique structure enables RNNs to effectively represent long-term dependencies in temporal data, which are well-suited for modelling systems with unknown dynamic orders \cite{ref7, ref8}. In \cite{ref9}, the regularization strategy was introduced into RNNs, which enhanced the network's robustness to noisy outliers and generalization performance. Unfortunately, these models rely on the error back-propagation (BP) algorithm to train weights and biases, which suffer from slow convergence, structure determination, and the sensitivity of the initial weights and learning parameters. Therefore, it is essential to develop advanced modelling techniques to address these problems.
 
In recent years, randomized methods for training NNs have received considerable attention due to their merits of fast learning speed and computational simplicity \cite{ref10, ref11, ref12, ref13}. Among them, regularized echo state networks (ESNs) have also been widely applied to tackle the ill-posedness in temporal data modelling \cite{ref15, ref16, ref17, ref18}. Nevertheless, ESNs lack a foundational understanding on the structure design and the random parameters assignment. With such a background, Wang and Li \cite{ref19} pioneered a novel randomized learner model, termed stochastic configuration networks (SCNs), which incorporated an innovative supervisory mechanism for assigning random parameters. This finding is significant for the advancement of randomized learning theory, and many promising results about nonlinear dynamic modelling have been reported. In \cite{ref20}, an enhanced version of SCN was presented by integrating L2 regularization, addressing issues related to structural settings and random parameter allocation. Furthermore, a regularized SCN was introduced in \cite{ref21} to solve the problems of outlier presence and multicollinearity in the input data, successfully mitigating concerns of overfitting. Wang et al. \cite{ref211} used weighted mean vectors to optimize the parameter selection and network structure of the regularized SCN, resulting in superior performance in terms of rapid convergence, network compactness, and predictive accuracy. Although SCNs and their regularization variants have shown progress in tackling industrial data analysis tasks, they are primarily designed for static data or based on the assumption of known system orders, posing challenges for modelling nonlinear systems with order uncertainty. 

Built on the SCN concept, we proposed the recurrent stochastic configuration networks (RSCNs) \cite{ref22}, where the reservoir structure is incrementally constructed and a specialized feedback matrix is established to ensure the universal approximation property. RSCNs offer several advantages, including the strong learning capabilities, computational simplicity, and the potential for implementing lightweight recurrent models. However, they struggle to identify useful order variables, resulting in information redundancy that complicates the extraction of true signals and increases the risk of overfitting. This paper presents an RSCN with hybrid regularization for problem solving. Firstly, the least absolute shrinkage and selection operator (LASSO) is used to extract the key order information from the temporal data. Then, an improved RSCN with L2 regularization is developed based on the selected order variables, as well as the residuals between the LASSO model output ${{\bf{Y}}_{LASSO}}$ and the target output ${\bf{T}}$. Finally, the output weights are dynamically adjusted according to the projection algorithm. Experimental results suggest that the proposed method can alleviate the ill-posed problem and perform favorably with sound generalization, indicating its strength in nonlinear dynamic modelling. The contributions of this paper can be summarized as follows:
\begin{itemize}
  \item [1)] 
A hybrid regularized modelling framework is introduced, which integrates LASSO and RSCN within a sequential ensemble learning strategy. This approach effectively captures relevant prior knowledge and reduces the data dimensionality. Moreover, the output compensation scheme trains the network by approximating the residuals ${\bf{\hat Y}} = {\bf{T}} - {{\bf{Y}}_{LASSO}}$, further enhancing the model's predictive accuracy. 
  \item [2)]
The random parameters are assigned in the light of an improved supervisory mechanism, which incorporates the regularization technique into the inequality constraint for generating the reservoir node. The L2 penalty term is introduced into the cost function of the original RSCN, and the output weights are evaluated by utilizing the regularized least squares method.
  \item [3)]
The theoretical analysis of the universal approximation property is provided. The echo state property of RSCNs is naturally inherited, and the convergence of the parameter updates based on the projection algorithm is also guaranteed. These results enable the built model to exhibit strong nonlinear processing capability and stability.
  \item [4)]
The impact of the regularization coefficient and reservoir size on the model performance is carefully considered, and two real-world industrial data predictive analyses are carried out. The findings illustrate the superiority of our proposed method in terms of structural compactness and generalization performance.
\end{itemize}

The remainder of this paper is organized as follows. Section II formulates the problem and reviews the related knowledge of RSCNs. Section III details the proposed hybrid regularized RSCN frameworks with the algorithm description and theoretical analysis. Section IV reports the experimental results. Finally, Section V concludes this paper.

\section{Related work}
In this section, two related works are introduced, including the problem formulation and recurrent stochastic configuration networks.

\subsection{Problem formulation}
Consider a discrete nonlinear dynamic system:
\begin{equation}
\label{eq001}
\begin{array}{l}
y\left( {n + 1} \right) = {f_{\rm{p}}}\left( {y\left( n \right), \ldots ,y\left( {n - {d_A} + 1} \right),{u_1}\left( n \right),} \right.\\
{\kern 1pt}  \ldots ,{u_1}\left( {n - d_B^1 + 1} \right),{u_2}\left( n \right), \ldots ,{u_2}\left( {n - d_B^2 + 1} \right),\\
\left. { \ldots ,{u_K}\left( n \right), \ldots ,{u_K}\left( {n - d_B^K + 1} \right)} \right) + e\left( n \right),
\end{array}
\end{equation}where $y$ is the plant output, $\left\{ {{u_1},{u_2}, \ldots ,{u_K}} \right\}$ is the input variable, $K$ is the input dimension, ${d_A},d_B^k$ are the systems orders for $y$ and ${u_k}$, $e$ is the process error, and ${f_{\rm{p}}}$ is a nonlinear function. Due to external disturbances and changes in the process environment, time-varying delays and unknown orders may occur in the system. These factors significantly affect the dynamic behavior of the system, resulting in a decline in the model's performance in both prediction and control \cite{ref23,ref24}. Specifically, time-varying delays may cause the system's response to become unstable, complicating the accurate capture of real-time changes, while unknown orders hinder the model's ability to identify and utilize the key variables. This uncertainty not only increases the complexity of modelling but also leads to decision-making errors, ultimately impacting the overall efficiency and reliability of the system.

This paper aims to address the challenge of modelling nonlinear systems with uncertain dynamic orders by extracting important order variable information as prior knowledge. This scheme helps researchers gain a clearer understanding of the system's dynamic characteristics, thereby enhancing the model's prediction performance. By identifying and analyzing these significant variables, the model can more effectively capture critical dynamic relationships within the system and reduce information redundancy. This not only aids in simplifying the model structure but also improves its adaptability in the face of complex environmental changes.

\subsection{Recurrent stochastic configuration networks}
This section briefly reviews our proposed RSCNs \cite{ref22}, which are a class of randomized learner models that hold the universal approximation property, and the random weights and biases are assigned in the light of a supervisory mechanism. Due to their advantages in neural network construction, such as less human intervention, adaptability of random parameters scope setting, fast learning, and handy implementation, RSCNs have demonstrated great potential in modelling nonlinear complex dynamics.

Given an RSCN model with $N$ reservoir nodes:
\begin{equation}
\label{eq1}
{{\bf{x}}_N}(n) = g({{\bf{W}}_{{\rm{in,}}N}}{\bf{u}}(n) + {{\bf{W}}_{{\mathop{\rm r}\nolimits} ,N}}{\bf{x}}(n - 1) + {{\bf{b}}_N}),
\end{equation}
\begin{equation}
\label{eq2}
{\bf{y}}(n) = {{\bf{W}}_{\rm{out}}}\left( {{{\bf{x}}_N}(n),{\bf{u}}(n)} \right),
\end{equation}
where $\mathbf{u}(n)\in {{\mathbb{R}}^{K}}$ is the input signal; $\mathbf{y}(n)$ is the model output; ${{\mathbf{x}}_{N}}(n)\in {{\mathbb{R}}^{N}}$ is the internal state of the reservoir; ${{\mathbf{W}}_{\text{in,}N}}\in {{\mathbb{R}}^{N\times K}},{{\mathbf{W}}_{\text{r,}N}}\in {{\mathbb{R}}^{N\times N}}$ represent the input and reservoir weights, respectively; ${{\mathbf{b}}_{N}}$ is the bias; $g$ is the activation function; ${{\mathbf{W}}_{\rm{out}}}\in {{\mathbb{R}}^{L\times \left( N+K \right)}}$ is the output weight; $K$ and $L$ are the dimensions of input and output. The model output can be obtained by $\mathbf{Y}=\left[ \mathbf{y}\left( 1 \right),\mathbf{y}\left( 2 \right),...,\mathbf{y}\left( {{n}_{max}} \right) \right]=\mathbf{W}_{\operatorname{out}}^{{}}{{\mathbf{X}}_{N}}$, where ${{n}_{\max }}$ is the number of training samples and ${{\mathbf{X}}_{N}}\text{=}\left[ \left( {{\mathbf{x}}_{N}}(1),\mathbf{u}(1) \right),\ldots ,\left( {{\mathbf{x}}_{N}}({{n}_{\max }}),\mathbf{u}({{n}_{\max }}) \right) \right]$.

Calculate the current error ${{e}_{N}}=\mathbf{Y}-\mathbf{T}$, where $\mathbf{T}=\left[ \mathbf{t}\left( 1 \right),\mathbf{t}\left( 2 \right), \right.$ $\left. \ldots ,\mathbf{t}\left( {{n}_{max}} \right) \right]$ is the desired output. If ${{e}_{N}}$ is larger than the preset error threshold $\varepsilon$, we need to assign the adding nodes based on the supervisory mechanism. Notably, to avoid overfitting, a supplementary condition is implemented to stop the addition of nodes and a step size ${N_{{\rm{step}}}}$ (${N_{{\rm{step}}}} < N$) is employed in the following early stopping criterion:
\begin{equation}
\label{eq002}
{\left\| {{e_{{\rm{val}},N - {N_{{\rm{step}}}}}}} \right\|_F} \le {\left\| {{e_{{\rm{val}},N - {N_{{\rm{step}}}} + 1}}} \right\|_F} \le  \ldots  \le {\left\| {{e_{{\rm{val}},N}}} \right\|_F},
\end{equation}
where ${{e}_{\text{val},N}}$ represents the validation residual error with $N$ nodes. If (\ref{eq002}) is satisfied, the number of reservoir nodes will be adjusted to $N-{{N}_{\text{step}}}$. The configuration process of the RSCN can be summarized as follows.

Step 1: Generate $\left[ \begin{matrix}
   w_{\operatorname{i}\operatorname{n}}^{N+1,1} & w_{\operatorname{i}\operatorname{n}}^{N+1,2} & \cdots  & w_{\operatorname{i}\operatorname{n}}^{N+1,K}  \\
\end{matrix} \right],$ ${{\mathbf{W}}_{\rm{r},N+1}}$, and ${{b}_{N+1}}$ stochastically in ${{G}_{\max }}$ times from the adjustable uniform distribution $\left[ -\lambda ,\lambda  \right]$, $\lambda \in \left\{ {{\lambda }_{\min }},{{\lambda }_{\min }}+\Delta \lambda ,...,{{\lambda }_{\max }} \right\}$. Construct a special structure of the reservoir weight matrix, that is,\vspace{-0.35cm}

\begin{small}
\begin{equation}
\label{eq5}
\begin{array}{l}
{{\bf{W}}_{\text{r,}2}}{\rm{ = }}\left[ {\begin{array}{*{20}{c}}
{w_{\mathop{\rm r}\nolimits} ^{1,1}}&0\\
{w_{\mathop{\rm r}\nolimits} ^{2,1}}&{w_{\mathop{\rm r}\nolimits} ^{2,2}}
\end{array}} \right],\\
{{\bf{W}}_{\text{r,}3}}{\rm{ = }}\left[ {\begin{array}{*{20}{c}}
{w_{\mathop{\rm r}\nolimits} ^{1,1}}&0&0\\
{w_{\mathop{\rm r}\nolimits} ^{2,1}}&{w_{\mathop{\rm r}\nolimits} ^{2,2}}&0\\
{w_{\mathop{\rm r}\nolimits} ^{3,1}}&{w_{\mathop{\rm r}\nolimits} ^{3,2}}&{w_{\mathop{\rm r}\nolimits} ^{3,3}}
\end{array}} \right],\\
 \ldots \\
{{\bf{W}}_{\text{r,}N{\rm{ + }}1}}{\rm{ = }}\left[ {\begin{array}{*{20}{c}}
{w_{\mathop{\rm r}\nolimits} ^{1,1}}&0& \cdots &0&0\\
{w_{\mathop{\rm r}\nolimits} ^{2,1}}&{w_{\mathop{\rm r}\nolimits} ^{2,2}}& \cdots &0&0\\
 \vdots & \vdots & \vdots & \vdots & \vdots \\
{w_{\mathop{\rm r}\nolimits} ^{N,1}}&{w_{\mathop{\rm r}\nolimits} ^{N,2}}& \cdots &{w_{\mathop{\rm r}\nolimits} ^{N,N}}&0\\
{w_{\mathop{\rm r}\nolimits} ^{N{\rm{ + 1,1}}}}&{w_{\mathop{\rm r}\nolimits} ^{N{\rm{ + 1,2}}}}& \cdots &{w_{\mathop{\rm r}\nolimits} ^{N{\rm{ + }}1,N}}&{w_{\mathop{\rm r}\nolimits} ^{N{\rm{ + }}1,N + 1}}
\end{array}} \right].
\end{array}
\end{equation}
\end{small}
\begin{remark}
The RSCN model holds the echo state property, where the current state of the model depends solely on the input data after a specific period, independent of the initial state. To ensure this property, ${{\bf{W}}_{{\rm{r}},N{\rm{ + }}1}}$ should be reset as
\begin{equation}
\label{eq501}
{{\bf{W}}_{{\rm{r}},N{\rm{ + }}1}} \leftarrow \frac{\alpha }{{\rho \left( {{{\bf{W}}_{{\rm{r}},N{\rm{ + 1}}}}} \right)}}{{\bf{W}}_{{\rm{r}},N{\rm{ + 1}}}},
\end{equation}
where $0 < \alpha  < {{{\rho _{\max }}\left( {{{\bf{W}}_{{\rm{r}},{N + 1}}}} \right)} \mathord{\left/
 {\vphantom {{{\rho _{\max }}\left( {{{\bf{W}}_{{\rm{r}},{N + 1}}}} \right)} {{\sigma _{\max }}\left( {{{\bf{W}}_{{\rm{r}},{N + 1}}}} \right)}}} \right.
 \kern-\nulldelimiterspace} {{\sigma _{\max }}\left( {{{\bf{W}}_{{\rm{r}},{N + 1}}}} \right)}}$ is the scaling factor, ${{\rho _{\max }}\left( {{{\bf{W}}_{{\rm{r}},{N + 1}}}} \right)}$ and ${{\sigma _{\max }}\left( {{{\bf{W}}_{{\rm{r}},{N + 1}}}} \right)}$ denote the largest eigenvalue and singular value of ${{\bf{W}}_{{\rm{r}},N{\rm{ + }}1}}$, respectively. For a detailed proof, one can refer to \cite{ref22}.
\end{remark}

The input weights are defined as
\begin{equation} \label{eq6}
\begin{array}{l}
{{\bf{W}}_{{\rm{in}},2}}{\rm{ = }}\left[ {\begin{array}{*{20}{l}}
{w_{{\mathop{\rm i}\nolimits} {\mathop{\rm n}\nolimits} }^{1,1}}&{w_{{\mathop{\rm i}\nolimits} {\mathop{\rm n}\nolimits} }^{1,2}}& \cdots &{w_{{\mathop{\rm i}\nolimits} {\mathop{\rm n}\nolimits} }^{1,K}}\\
{w_{{\mathop{\rm i}\nolimits} {\mathop{\rm n}\nolimits} }^{2,1}}&{w_{{\mathop{\rm i}\nolimits} {\mathop{\rm n}\nolimits} }^{2,2}}& \cdots &{w_{{\mathop{\rm i}\nolimits} {\mathop{\rm n}\nolimits} }^{2,K}}
\end{array}} \right],\\
 \ldots \\
{{\bf{W}}_{{\rm{in}},N + 1}}{\rm{ = }}\left[ {\begin{array}{*{20}{c}}
{w_{{\mathop{\rm i}\nolimits} {\mathop{\rm n}\nolimits} }^{1,1}}&{w_{{\mathop{\rm i}\nolimits} {\mathop{\rm n}\nolimits} }^{1,2}}& \cdots &{w_{{\mathop{\rm i}\nolimits} {\mathop{\rm n}\nolimits} }^{1,K}}\\
{w_{{\mathop{\rm i}\nolimits} {\mathop{\rm n}\nolimits} }^{2,1}}&{w_{{\mathop{\rm i}\nolimits} {\mathop{\rm n}\nolimits} }^{2,2}}& \cdots &{w_{{\mathop{\rm i}\nolimits} {\mathop{\rm n}\nolimits} }^{2,K}}\\
 \vdots & \vdots & \vdots & \vdots \\
{w_{{\mathop{\rm i}\nolimits} {\mathop{\rm n}\nolimits} }^{N,1}}&{w_{{\mathop{\rm i}\nolimits} {\mathop{\rm n}\nolimits} }^{N,2}}& \cdots &{w_{{\mathop{\rm i}\nolimits} {\mathop{\rm n}\nolimits} }^{N,K}}\\
{w_{{\mathop{\rm i}\nolimits} {\mathop{\rm n}\nolimits} }^{N + 1,1}}&{w_{{\mathop{\rm i}\nolimits} {\mathop{\rm n}\nolimits} }^{N + 1,2}}& \cdots &{w_{{\mathop{\rm i}\nolimits} {\mathop{\rm n}\nolimits} }^{N + 1,K}}
\end{array}} \right],
\end{array}
\end{equation}
and ${{\mathbf{b}}_{2}}={{\left[ {{b}_{1}},{{b}_{2}} \right]}^{\top }},\ldots ,{{\mathbf{b}}_{N+1}}={{\left[ {{b}_{1}},\ldots, {{b}_{N+1}} \right]}^{\top }}$.

Step 2: Seek the random basis function ${g_{N{\rm{ + }}1}}$ satisfying the following inequality constraint with an increasing factor ${{r}_{i}}$, $i=1,2,\ldots ,t$, where $0<{{r}_{1}}<{{r}_{2}}<\ldots <{{r}_{t}}<1$.
\begin{equation}
\label{eq7}
{\left\langle {e_{N,q}^{},{g_{N{\rm{ + }}1}}} \right\rangle ^2} \ge b_g^2(1 - {r_i} - {\mu _{N + 1}})\left\| {{e_{N,q}}} \right\|_{}^2,q = 1,2,...L,
\end{equation}
where ${{e}_{N}}={{\left[ {{e}_{N,1}},{{e}_{N,2}},\ldots ,{{e}_{N,L}} \right]}^{\top }}$ is the residual error, $\left\{ {{\mu }_{N+1}} \right\}$ is a non-negative real sequence satisfying $\underset{N\to \infty }{\mathop{\lim }}\,{{\mu }_{N+1}}=0$, ${{\mu }_{N+1}}\le (1-r)$, and $0<\left\| g \right\|<{{b}_{g}}$, ${b_g} \in {{\mathbb{R}}^+ }$.

Step 3: Define a set of variables $\left[ {{\xi }_{N+1,1}},...,{{\xi }_{N+1,L}} \right]$, that is,
\begin{equation}
\label{eq8}
{\xi _{N + 1,q}} = \frac{{{{\left( {e_{N,q}^ \top {g_{N + 1}}} \right)}^2}}}{{g_{N + 1}^ \top {g_{N + 1}}}} - \left( {1 - {\mu _{N + 1}} - r} \right)e_{N,q}^ \top e_{N,q}^{}.
\end{equation}
Select the node with the largest ${{\xi }_{N+1}}=\sum\limits_{q=1}^{L}{{{\xi }_{N+1,q}}}$ as the optimal adding node.

Step 4: Evaluate the output weights based on the least square method, that is,
\begin{equation}
\label{eq9}
\begin{array}{l}
{\bf{W}}_{{\rm{out}},N{\rm{ + }}1}^{}{\rm{ = }}\left[ {{\bf{w}}_{{\rm{out}},1}^{},{\bf{w}}_{{\rm{out}},2}^{},...,{\bf{w}}_{{\rm{out}},N + 1 + K}^{}} \right]\\
{\kern 1pt} {\kern 1pt} {\kern 1pt} {\kern 1pt} {\kern 1pt} {\kern 1pt} {\kern 1pt} {\kern 1pt} {\kern 1pt} {\kern 1pt} {\kern 1pt} {\kern 1pt} {\kern 1pt} {\kern 1pt} {\kern 1pt} {\kern 1pt} {\kern 1pt} {\kern 1pt} {\kern 1pt} {\kern 1pt} {\kern 1pt} {\kern 1pt} {\kern 1pt} {\kern 1pt} {\kern 1pt} {\kern 1pt} {\kern 1pt} {\kern 1pt} {\kern 1pt} {\kern 1pt} {\kern 1pt} {\kern 1pt}{\kern 1pt} {\kern 1pt} {\kern 1pt} {\kern 1pt} {\kern 1pt} {\kern 1pt} {\kern 1pt} {\kern 1pt} {\kern 1pt}  = \mathop {\arg \min }\limits_{{\bf{W}}_{{\rm{out}}}^{}} \left\| {{\bf{T}} - {\bf{W}}_{{\rm{out}}}^{}{{\bf{X}}_{N + 1}}} \right\|_{}^2,
\end{array}
\end{equation}
where \small ${{\mathbf{X}}_{N+1}}\text{=}\left[ \left( {{\mathbf{x}}_{N+1}}\left( 1 \right),\mathbf{u}\left( 1 \right) \right),\ldots ,\left( {{\mathbf{x}}_{N+1}}\left( {{n}_{\max }} \right),\mathbf{u}\left( {{n}_{\max }} \right) \right) \right]$. \normalsize

Step 5: Calculate the residual error ${{e}_{N+1}}$ and update ${{e}_{0}}:={{e}_{N+1}}$, $N=N+1$. If ${{\left\| {{e}_{0}} \right\|}_{F}}\le \varepsilon $, $N\ge {{N}_{\max }}$ or (\ref{eq002}) is met, we complete the configuration, where ${{N}_{\max }}$ is the maximum size of the reservoir. Otherwise, repeat steps 1-4 until satisfying the termination conditions. 

At last, we have $\underset{N\to \infty }{\mathop{\lim }}\,{{\left\| {{e}_{N}} \right\|}_{F}}=0$. For more details, readers can refer to \cite{ref22}.

\section{Regularized recurrent stochastic configuration networks}
This section details the construction process of our proposed hybrid regularized RSCN framework and presents the theoretical analysis on the universal approximation property. 

\subsection{Variables selection with LASSO}
LASSO regression is an effective method for feature selection and modelling, particularly for high-dimensional data \cite{ref25,ref26,ref27}. Consider a linear regression model, the LASSO regularization solution can be formulated by
\begin{equation} \label{eq121}
\mathop {\min }\limits_{\bf{W}} \left\| {{{\bf{W}}_{LASSO}}{\bf{U}} - {\bf{T}}} \right\|_2^2 + {C_L}{\left\| {{\bf{W}}_{LASSO}^{}} \right\|_1},
\end{equation}
where ${{\bf{W}}_{LASSO}} = \left[ {{{\bf{W}}_1},{{\bf{W}}_2}, \ldots ,{{\bf{W}}_L}} \right]$ is the regression coefficient, ${\bf{U}}$ is the input, and ${C_L}$ is a regularization coefficient. $\left\|  \bullet  \right\|_1$ and $\left\|  \bullet  \right\|_2$ represent the L1-norm and L2-norm. By adjusting ${C_L}$, the regression coefficients of redundant or irrelevant features can be compressed to zero, resulting in a sparser linear regression model that facilitates feature extraction. Consequently, only the order variables corresponding to non-zero regression coefficients are retained.

Calculate the output of the LASSO model ${{\bf{Y}}_{LASSO}} = {{\bf{W}}_{LASSO}}{\bf{U}}$, and obtain the residuals between ${{\bf{Y}}_{LASSO}}$ and the the target output, that is,
\begin{equation} \label{eq122}
{\bf{\hat Y}} = {\bf{T}} - {{\bf{Y}}_{LASSO}}.
\end{equation}
Then, use the selected order variables denoted as ${{\bf{U}}_{\rm{R}}}$ to construct the regularized RSCN model and continuously approximate ${\bf{\hat Y}}$. This order identification and output compensation scheme not only enhances the model's simplicity but also improves its predictive capability.

\begin{remark}
Feature selection and order identification are two distinct concepts in data analysis and modelling, both of which aim to extract useful information from the raw data to improve the model performance. Feature selection primarily focuses on deriving representative features from the data and is widely applied in areas such as image processing and natural language processing. In contrast, order identification emphasizes the recognition of important time delays or order variables within a system, commonly encountered in dynamic system modelling and time series forecasting. This paper seeks to provide prior order information for system modelling, thereby improving the interpretation of temporal data through the clear identification of key order variables and enhancing the network's ability to capture complex dynamic behaviors.
\end{remark}

\subsection{Regularized recurrent stochastic configuration networks}
Distinguished from the original RSCN, the proposed method integrates regularization techniques into the supervision mechanism and introduces an L2 penalty term to the loss function to limit the magnitude of the output weights. This design effectively manages the complexity of the model, further preventing overfitting to the training data and improving the model's generalization capability on the newly arriving data.

Given the input ${{\bf{U}}_{\rm{R}}}{\rm{ = }}\left[ {{{\bf{u}}_{\rm{R}}}(1), \ldots ,{{\bf{u}}_{\rm{R}}}({n_{\max }})} \right]$ and desired output ${\bf{\hat Y}}$, an initial model with $N$ nodes can be constructed. If the terminal conditions are not met, we need to generate a new basis function ${{g}_{N+1}}$ under the supervisory mechanism with regularization.\\

\textbf{Theorem 1.} Suppose that span$\left( \Gamma  \right)$ is dense in L2 space. Define ${{\rho }_{g}}={{{\left( \left\| g \right\|_{{}}^{2}+C \right)}^{2}}}/{\left( \left\| g \right\|_{{}}^{2}+2C \right)}\;$,\\ ${{\rho }_{{{b}_{g}}}}={{{\left( b_{g}^{2}+C \right)}^{2}}}/{\left( b_{g}^{2}+2C \right)}\;$, $0 < C < 1$, for $b_{g}\in {{\mathbb{R}}^{+}}$, $\forall {g}\in \Gamma ,0<{{\left\| {{g}} \right\|}}<b_{g}$, and $0<{{\rho }_{g}}<{{\rho }_{{{b}_{g}}}}$. Given $0<r<1$ and a nonnegative real sequence $\left\{ {{\mu }_{N+1}} \right\}$ satisfying $\underset{N\to \infty }{\mathop{\lim }}\,{{\mu }_{N+1}}=0$, and ${{\mu }_{N+1}}\le (1-r)$. For $N=1,2...$, $q=1,2,...,L$, define
\begin{equation} \label{eq15}
{\delta _{N{\rm{ + 1}}}} = \sum\limits_{q = 1}^L {\delta _{N{\rm{ + }}1,q}^{}} {\kern 1pt} {\kern 1pt} {\kern 1pt} {\kern 1pt} ,{\kern 1pt} {\kern 1pt} {\kern 1pt} \delta _{N{\rm{ + }}1,q}^{} = (1 - r - {\mu _{N{\rm{ + }}1}})\left\| {e_{N,q}^{}} \right\|_{}^2.
\end{equation}
If ${{g}_{N+1}}$ satisfies the following inequality constraint:
\begin{equation} \label{eq16}
{\left\langle {e_{N,q}^{},g_{N{\rm{ + }}1}^{}} \right\rangle ^2} \ge \frac{{{{\left( {b_g^2 + C} \right)}^2}}}{{\left( {b_g^2 + 2C} \right)}}\delta _{N{\rm{ + }}1,q}^{},q = 1,2, \ldots L,
\end{equation}
and the output weight is constructively evaluated by
\begin{equation} \label{eq17}
{\bf{w}}_{{\mathop{\rm out}\nolimits} ,N + 1,q}^{} = \frac{{\left\langle {e_{N,q}^{},g_{N{\rm{ + }}1}^{}} \right\rangle }}{{\left\| {g_{N{\rm{ + }}1}^{}} \right\|_{}^2 + C}},
\end{equation}
we have ${{\lim }_{N\to \infty }}\left\| e_{N\text{+}1} \right\|=0$.\\

\textbf{Proof.} Note that
\begin{equation} \label{eq18}
\begin{array}{l}
\left\| {e_{N{\rm{ + }}1}^{}} \right\|_{}^2 - \left( {r + {\mu _{N{\rm{ + }}1}}} \right)\left\| {e_N^{}} \right\|_{}^2\\
 = \sum\limits_{q = 1}^L {\left\langle {e_{N,q}^{} - {\bf{w}}_{{\mathop{\rm out}\nolimits} ,N + 1,q}^{}g_{N + 1}^{},} \right.} {\kern 1pt} {\kern 1pt} \left. {{\kern 1pt} {\kern 1pt} e_{N,q}^{} - {\bf{w}}_{{\mathop{\rm out}\nolimits} ,N + 1,q}^{}g_{N + 1}^{}} \right\rangle \\
{\kern 1pt} {\kern 1pt} {\kern 1pt} {\kern 1pt} {\kern 1pt} {\kern 1pt} {\kern 1pt} {\kern 1pt} {\kern 1pt} {\kern 1pt} {\kern 1pt} {\kern 1pt} {\kern 1pt} {\kern 1pt} {\kern 1pt} {\kern 1pt} {\kern 1pt} {\kern 1pt} {\kern 1pt} {\kern 1pt}  - \sum\limits_{q = 1}^L {\left( {r + {\mu _{N{\rm{ + }}1}}} \right)\left\langle {e_{N,q}^{},e_{N,q}^{}} \right\rangle } \\
 = \left( {1 - r - {\mu _{N{\rm{ + }}1}}} \right)\left\| {e_N^{}} \right\|_{}^2 - \sum\limits_{q = 1}^L {\left( {2\left\langle {e_{N,q}^{},{\bf{w}}_{{\mathop{\rm out}\nolimits} ,N + 1,q}^{}g_{N + 1}^{}} \right\rangle } \right.} \\
{\kern 1pt} {\kern 1pt} {\kern 1pt} {\kern 1pt} {\kern 1pt} {\kern 1pt} {\kern 1pt} {\kern 1pt} {\kern 1pt} {\kern 1pt} {\kern 1pt} {\kern 1pt} {\kern 1pt} {\kern 1pt} {\kern 1pt} {\kern 1pt} {\kern 1pt} {\kern 1pt} {\kern 1pt} \left. { - \left\langle {{\bf{w}}_{{\mathop{\rm out}\nolimits} ,N + 1,q}^{}g_{N + 1}^{},{\bf{w}}_{{\mathop{\rm out}\nolimits} ,N + 1,q}^{}g_{N + 1}^{}} \right\rangle } \right)\\
= \left( {1 - r - {\mu _{N{\rm{ + }}1}}} \right)\left\| {e_N^{}} \right\|_{}^2\\
{\kern 1pt} {\kern 1pt} {\kern 1pt} {\kern 1pt} {\kern 1pt} {\kern 1pt} {\kern 1pt} {\kern 1pt} {\kern 1pt} {\kern 1pt} {\kern 1pt} {\kern 1pt} {\kern 1pt} {\kern 1pt} {\kern 1pt} {\kern 1pt} {\kern 1pt}  - \sum\limits_{q = 1}^L {\left( {\left\langle {{\bf{w}}_{{\mathop{\rm out}\nolimits} ,N + 1,q}^{}g_{N + 1}^{},{\bf{w}}_{{\mathop{\rm out}\nolimits} ,N + 1,q}^{}g_{N + 1}^{}} \right\rangle } \right.} \\
\left. {{\kern 1pt} {\kern 1pt} {\kern 1pt} {\kern 1pt} {\kern 1pt} {\kern 1pt} {\kern 1pt} {\kern 1pt} {\kern 1pt} {\kern 1pt} {\kern 1pt} {\kern 1pt} {\kern 1pt} {\kern 1pt} {\kern 1pt} {\kern 1pt} {\kern 1pt} {\kern 1pt}  + 2C\left\langle {{\bf{w}}_{{\mathop{\rm out}\nolimits} ,N + 1,q}^{},{\bf{w}}_{{\mathop{\rm out}\nolimits} ,N + 1,q}^{}} \right\rangle } \right)\\
 = \left( {1 - r - {\mu _{N{\rm{ + }}1}}} \right)\left\| {e_N^{}} \right\|_{}^2 - \frac{{\sum\limits_{q = 1}^L {{{\left\langle {e_{N,q}^{},g_{N + 1}^{}} \right\rangle }^2}} }}{{{{{{\left( {\left\| {g_{N{\rm{ + }}1}^{}} \right\|_{}^2{\rm{ + }}C} \right)}^2}} \mathord{\left/
 {\vphantom {{{{\left( {\left\| {g_{N{\rm{ + }}1}^{}} \right\|_{}^2{\rm{ + }}C} \right)}^2}} {\left( {\left\| {g_{N{\rm{ + }}1}^{}} \right\|_{}^2{\rm{ + 2}}C} \right)}}} \right.
 \kern-\nulldelimiterspace} {\left( {\left\| {g_{N{\rm{ + }}1}^{}} \right\|_{}^2{\rm{ + 2}}C} \right)}}}}\\
 = \delta _{N{\rm{ + }}1}^{} - \frac{{\sum\limits_{q = 1}^L {{{\left\langle {e_{N,q}^{},g_{N + 1}^{}} \right\rangle }^2}} }}{{{{{{\left( {\left\| {g_{N{\rm{ + }}1}^{}} \right\|_{}^2{\rm{ + }}C} \right)}^2}} \mathord{\left/
 {\vphantom {{{{\left( {\left\| {g_{N{\rm{ + }}1}^{}} \right\|_{}^2{\rm{ + }}C} \right)}^2}} {\left( {\left\| {g_{N{\rm{ + }}1}^{}} \right\|_{}^2{\rm{ + 2}}C} \right)}}} \right.
 \kern-\nulldelimiterspace} {\left( {\left\| {g_{N{\rm{ + }}1}^{}} \right\|_{}^2{\rm{ + 2}}C} \right)}}}}\\
 \le \delta _{N{\rm{ + }}1}^{} - \frac{{\sum\limits_{q = 1}^L {{{\left\langle {e_{N,q}^{},g_{N + 1}^{}} \right\rangle }^2}} }}{{{{{{\left( {b_g^2{\rm{ + }}C} \right)}^2}} \mathord{\left/
 {\vphantom {{{{\left( {b_g^2{\rm{ + }}C} \right)}^2}} {\left( {b_g^2{\rm{ + 2}}C} \right)}}} \right.
 \kern-\nulldelimiterspace} {\left( {b_g^2{\rm{ + 2}}C} \right)}}}}.
\end{array}
\end{equation}
According to (\ref{eq16}), we have $\left\| e_{N\text{+}1}^{{}} \right\|_{{}}^{2}-\left( r+{{\mu }_{N\text{+}1}} \right)\left\| e_{N}^{{}} \right\|_{{}}^{2}\le 0$. Notably, the global regularized least squares method is used to calculate the output weight in our proposed algorithm. The output weight ${{\mathbf{W}}_{\rm{out}}}$ can be evaluated by
\begin{equation} \label{eq13}
\mathop {\min }\limits_{{\bf{W}}_{{\mathop{\rm out}\nolimits} }^{}} \left\| {{\bf{W}}_{{\mathop{\rm out}\nolimits} }^{}{{{\bf{\hat X}}}_{N + 1}} - {\bf{\hat Y}}} \right\|_2^2 + C\left\| {{\bf{W}}_{{\mathop{\rm out}\nolimits} }^{}} \right\|_2^2,
\end{equation}
\begin{equation} \label{eq14}
{\bf{W}}_{{\mathop{\rm out}\nolimits} }^ \top  = {\left( {{{{\bf{\hat X}}}_{N + 1}}{\bf{\hat X}}_{N + 1}^ \top  + C{\bf{I}}} \right)^{ - 1}}{{\bf{\hat X}}_{N + 1}}{{\bf{\hat Y}}^ \top },
\end{equation}
where \footnotesize ${{\bf{\hat X}}_{N + 1}}{\rm{ = }}\left[ {\left( {{g_{N + 1}}\left( 1 \right);{{\bf{u}}_{\rm{R}}}\left( 1 \right)} \right), \ldots ,\left( {{g_{N + 1}}\left( {{n_{\max }}} \right);} \right.} \right.$$\left. {\left. {{{\bf{u}}_{\rm{R}}}\left( {{n_{\max }}} \right)} \right)} \right]$, \normalsize $C$ is the regularization coefficient, $\mathbf{I}$ is an identity matrix, and the model output is ${{\bf{Y}}_{{\rm{RSCN - L2}}}} = {{\bf{W}}_{{\rm{out}}}}{{\bf{\hat X}}_{N + 1}}$.

Then, we can obtain the optimum output weight $\mathbf{W}_{\text{out},N+1}^{{}}\text{=}\left[ \mathbf{w}_{\text{out},1}^{*},\mathbf{w}_{\text{out},2}^{*}, \right.\left. ...,\mathbf{w}_{\text{out},N+K}^{*} \right]$ and residual error $\tilde{e}_{N\text{+}1}^{{}}$ 
 by combining (\ref{eq13}) and (\ref{eq14}). It is easily inferred that $\left\| \tilde{e}_{N\text{+}1}^{{}} \right\|_{{}}^{2}\le \left\| e_{N\text{+}1}^{{}} \right\|_{{}}^{2}$. Thus, the following inequality holds:
\begin{equation} \label{eq19}
\left\| {\tilde e_{N{\rm{ + }}1}^{}} \right\|_{}^2 \le r\left\| {e_N^{}} \right\|_{}^2 + {\gamma _{N + 1}},
\end{equation} where ${{\gamma _{N + 1}} = {\mu _{N{\rm{ + }}1}}\left\| {e_N^{}} \right\|_{}^2 \ge 0}$.
Note that $\mathop {\lim }\limits_{N \to \infty } {\gamma _{N + 1}} = 0$. From (\ref{eq19}), we have $\underset{N\to \infty }{\mathop{\lim }}\,\left\| {{{\tilde{e}}}_{N\text{+}1}} \right\|_{{}}^{2}=0$, which completes the proof.

\subsection{Online adjustment of learning parameters}
To respond to the rapid changes in the system, it is necessary to adjust the model's learning parameters based on real-time data. This subsection elaborates on the online update of network parameters based on the projection algorithm \cite{ref28}.

Given a regularized RSCN model, at $n$-th step, let ${\bf{\hat g}}(n) = \left( {{{\bf{x}}_N}(n),{{\bf{u}}_{\rm{R}}}(n)} \right)$, where ${{{\bf{x}}_N}(n)}$ denote the internal state of the reservoir with $N$ nodes. Determine ${{\mathbf{W}}_{{\rm out}}}\left( n \right)$ to minimize the following cost function:
\begin{equation} \label{eq191}
\begin{array}{*{20}{c}}
{J = \frac{1}{2}{{\left\| {{{{\mathbf{W}}}_{{\rm out}}}\left( n \right) - {{{\mathbf{W}}}_{{\rm out}}}\left( {n - 1} \right)} \right\|}^2}}\\
{s.t.{\kern 1pt} {\kern 1pt} {\kern 1pt} {\kern 1pt} {\kern 1pt} {\kern 1pt} {\kern 1pt} {\kern 1pt} {\kern 1pt} {\kern 1pt} {\bf{y}}\left( n \right) = {{{\mathbf{W}}}_{{\rm out}}}\left( {n - 1} \right){\bf{\hat g}}(n)}
\end{array}.
\end{equation}
By incorporating the Lagrange multiplier ${{\lambda }_{\text{p}}}$, we can obtain
\begin{equation} \label{eq192}
\begin{array}{l}
{J_e} = \frac{1}{2}{\left\| {{{{\mathbf{W}}}_{\rm out}}\left( n \right) - {{{\mathbf{W}}}_{\rm out}}\left( {n - 1} \right)} \right\|^2}\\
{\kern 1pt} {\kern 1pt} {\kern 1pt} {\kern 1pt} {\kern 1pt} {\kern 1pt} {\kern 1pt} {\kern 1pt} {\kern 1pt} {\kern 1pt} {\kern 1pt} {\kern 1pt} {\kern 1pt} {\kern 1pt} {\kern 1pt} {\kern 1pt}  + {\lambda _{\rm{p}}}\left[ {{\bf{y}}\left( n \right) - {{{\mathbf{W}}}_{\rm out}}\left( {n - 1} \right){\bf{\hat g}}(n)} \right].
\end{array}
\end{equation}
Taking the derivative of (\ref{eq192}) with respect to ${{\mathbf{W}}_{{\rm out}}}\left( n \right)$ and ${{\lambda }_{\text{p}}}$, yields
\begin{equation} \label{eq193}
\left\{ {\begin{array}{*{20}{c}}
{{{{\mathbf{W}}}_{{\rm out}}}\left( n \right) - {{{\mathbf{W}}}_{{\rm out}}}\left( {n - 1} \right) - {\lambda _{\rm{p}}}{{{\bf{\hat g}}}^ \top }(n) = 0}\\
{{\bf{y}}\left( n \right) - {{{\mathbf{W}}}_{{\rm out}}}\left( {n - 1} \right){\bf{\hat g}}(n) = 0}
\end{array}} \right.,
\end{equation}
\begin{equation} \label{eq194}
{\lambda _{\rm{p}}} = \frac{{{\bf{y}}\left( n \right) - {{{\mathbf{W}}}_{{\rm out}}}\left( {n - 1} \right){\bf{\hat g}}(n)}}{{{\bf{\hat g}}{{(n)}^ \top }{\bf{\hat g}}(n)}}.
\end{equation}
Substituting (\ref{eq194}) into (\ref{eq193}), we have
\begin{equation} \label{eq195}
\begin{array}{l}
{{{\mathbf{W}}}_{\rm out}}(n) = {{{\mathbf{W}}}_{\rm out}}(n - 1)  \\
{\kern 1pt} {\kern 1pt} {\kern 1pt} {\kern 1pt} {\kern 1pt} {\kern 1pt} {\kern 1pt} {\kern 1pt} {\kern 1pt} {\kern 1pt} {\kern 1pt} {\kern 1pt} {\kern 1pt} {\kern 1pt} {\kern 1pt} {\kern 1pt} {\kern 1pt} {\kern 1pt} {\kern 1pt} {\kern 1pt} +\frac{{{\bf{\hat g}}{{(n)}^ \top }}}{{{\bf{\hat g}}{{(n)}^ \top }{\bf{\hat g}}(n)}}\left( {{\bf{y}}\left( n \right) - {{{\mathbf{W}}}_{\rm out}}\left( {n - 1} \right){\bf{\hat g}}(n)} \right).
\end{array}
\end{equation}

\begin{remark}
This online adjustment enables the network to quickly adapt to dynamic unknown data. Moreover, we have analyzed the stability and convergence of the proposed scheme and presented an enhanced condition to further improve the model's stability in \cite{ref22}.
\end{remark}
\begin{figure}[ht]
\vspace{-0.3cm}
	\begin{center}
		\includegraphics[width=9cm]{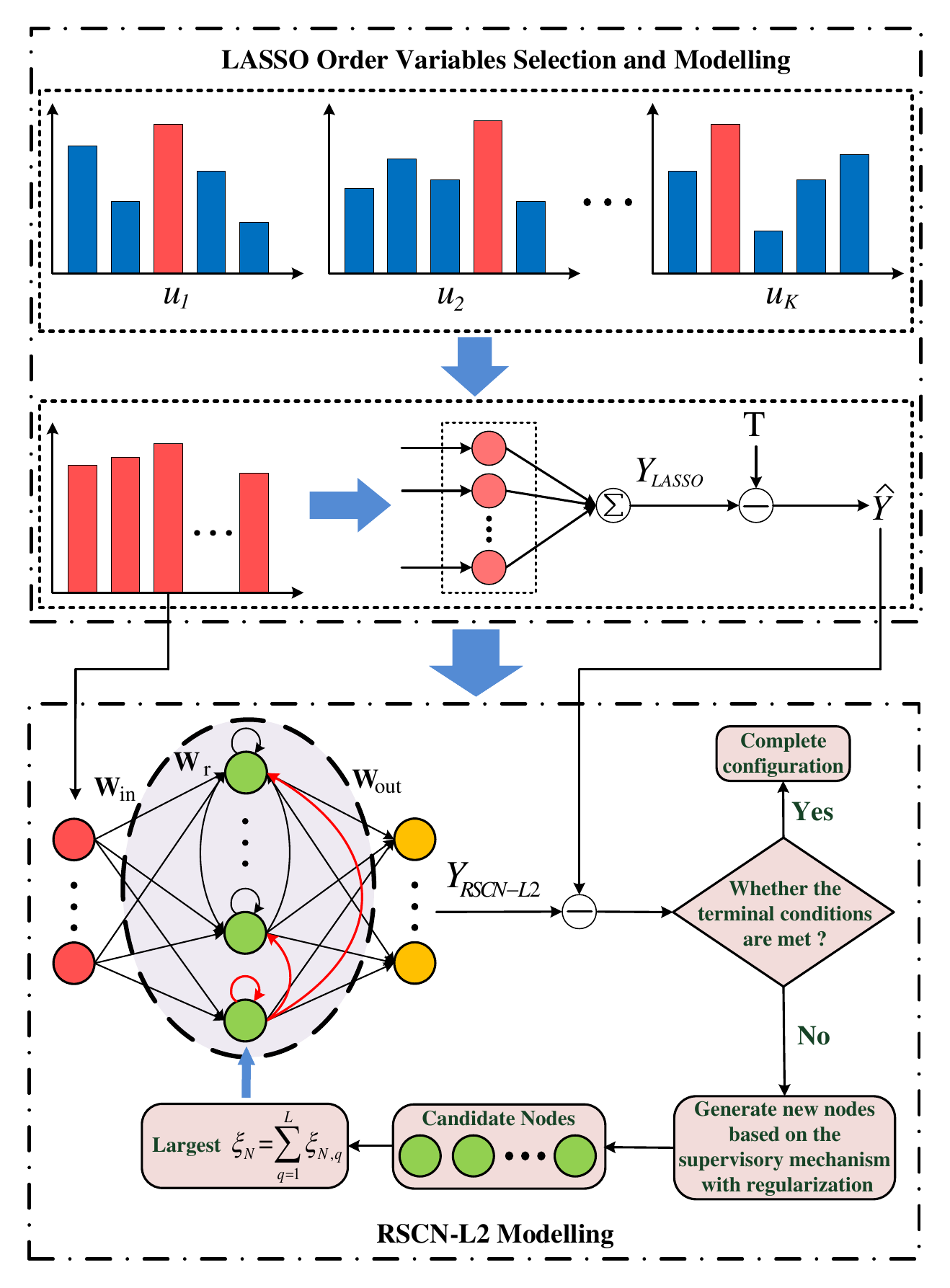}
		\caption{Architectures of the hybrid regularized RSCN.}
		\label{fig1}
	\end{center}
    \vspace{-0.5cm}
\end{figure}

\begin{algorithm}[t]\footnotesize
    \caption{RSC with hybrid regularization}\label{algo1}	
    \KwIn{Training dataset $\left\{ {{\bf{U}},{\bf{T}}} \right\}$, validation dataset $\left\{ {{{\bf{U}}_{{\rm{val}}}},{{\bf{T}}_{{\rm{val}}}}} \right\}$, the size of the reservoir $N$, the maximum number of reservoir size $N_{\max }$, the step size ${N_{{\rm{step}}}}$, training error threshold $\varepsilon $, random parameter scalars ${\bf{\gamma }}{\rm{ = }}\left\{ {{\lambda _1},{\lambda _2},...,{\lambda _{\max }}} \right\}$, the maximum number of stochastic configurations ${{G}_{\max }}$, and the regularization coefficient $C$.}
    \KwOut{Hybrid regularized RSCN}
    Extract the key order variables and establish a linear regression model based on (\ref{eq121}) and (\ref{eq122});\\
    Obtain the elected order variables ${{\bf{U}}_{\rm{R}}}$ and ${{\bf{U}}_{{\rm{R,val}}}}$, and the residuals ${\bf{\hat Y}}$ and ${{\bf{\hat Y}}_{{\rm{val}}}}$ for the two datasets, respectively;\\
    Randomly assign $\mathbf{W}_{\text{in,}N}$, $\mathbf{W}_{\text{r,}N}$, and $\mathbf{b}_{N}$ according to the sparsity of the reservoir from $\left[ -\lambda ,\lambda  \right]$. Calculate the model output and current error ${{e}_{N}}$. Set the initial residual error ${{e}_{0}}:={{e}_{N}}$, $0<r<1$,  $\mathbf{\Omega },\mathbf{D}:=\left[ {\kern 1pt}{\kern 1pt} \right]$;\\
    \While {$N<{{N}_{\max }}$ AND ${{\left\| {{e}_{0}} \right\|}_{F}}>\varepsilon $}{
        \If{the early stopping criterion in (\ref{eq002}) is not satisfied}{
        \For{$\lambda \in \mathbf{\gamma }$,}{
            \For{$i=1,2,\ldots ,{{G}_{\max }}$,}{
                Randomly assign ${\bf{W}}_{{\rm{in,}}N + 1}$, ${{\bf{b}}_{N + 1}}$, and ${\bf{w}}_{\rm{r}}^{N + 1} = \left[ {w_{\rm{r}}^{N + 1,1}, \ldots ,w_{\rm{r}}^{N + 1,N + 1}} \right]$ from $\left[ -\lambda ,\lambda  \right]$;\\
                Construct ${{\bf{W}}_{{\rm{r}},N + 1}}$ according to (\ref{eq5});\\
                Calculate the basis function ${g_{N + 1}}$ and $\xi _{N + 1,q}^ * $;\\
                Set ${\mu _{N + 1}} = {{\left( {1 - r} \right)} \mathord{\left/
                    {\vphantom {{\left( {1 - r} \right)} {N + 1}}} \right.
                    \kern-\nulldelimiterspace} {N + 1}}$;\\
                \If{$\min \left\{ {\xi _{N{\rm{ + }}1,1}^ * ,\xi _{N{\rm{ + }}1,2}^ * ,...,\xi _{N{\rm{ + }}1,L}^ * } \right\} \ge 0$}{
                    Save ${\bf{W}}_{{\rm{in,}}N + 1}$, ${{\bf{b}}_{N + 1}}$, and ${{\bf{W}}_{{\rm{r}},N + 1}}$ in $\mathbf{D}$, and $\xi _{N + 1}^ * {\rm{ = }}\sum\limits_{q = 1}^L {\xi _{N + 1,q}^ * } $ in $\mathbf{\Omega }$;\\
                    \Else{Go back to \textbf{Step 7}}\
                }    
            }
            \If{$\mathbf{D}$ is not empty}{
               Find ${\bf{W}}_{{\rm{in,}}N + 1}^ * $, ${\bf{b}}_{N + 1}^ * $, and ${\bf{W}}_{{\rm{r}},N + 1}^ * $ that maximize $\xi _{N + 1}^ * $, and get ${{\bf{\hat X}}_{N + 1}}$;\\
               \textbf{Break} (go to \textbf{Step 29});\\
                \Else{Randomly take $\tau \in \left( 0,1-r \right)$, update $r=r+\tau $, and return to \textbf{Step 7}}\
                }
        }
        Calculate ${\bf{W}}_{{\mathop{\rm out}\nolimits} }^ * $ and ${{\bf{Y}}_{{\rm{RSCN - L2}}}}$ based on (\ref{eq13}) and (\ref{eq14});\\
        Calculate ${e_{N + 1}}$ and ${e_{{\rm{val}},N + 1}}$;\\
        Update ${{e}_{0}}:={{e}_{N+1}}$, and $N=N+1$; \\ 
        \Else{Complete the configuration and set the reservoir size to $N:=N-{{N}_{\text{step}}}$;\\
            \textbf{Break} (go to \textbf{Step 38})}
        }
    }
\textbf{Return} ${\bf{W}}_{{\rm{in,}}N + 1}^ * $, ${\bf{b}}_{N + 1}^ * $, ${\bf{W}}_{{\rm{r}},N + 1}^ * $, and $\mathbf{W}_{\operatorname{out}}^{*}$.
\end{algorithm}

\subsection{Algorithm description}
To extract effective order variables and solve ill-posed problems in nonlinear dynamic modelling, this paper presents a hybrid regularized RSCN framework. An improved supervisory mechanism with regularization is introduced to incrementally construct the reservoir topology. The implementation of the proposed approach is illustrated in Fig. \ref{fig1}, and the details are as follows.

Step 1: Identify the significant order variables ${{\bf{U}}_{\rm{R}}}$ and calculate the corresponding residuals based on (\ref{eq121}) and (\ref{eq122}). Employ them as the input and target output for the regularized RSCN, respectively.

Step 2: Build an initial model with $N$ nodes and compute the residual error ${{e}_{N}}$. If the terminal conditions are not satisfied, we need to configure the new nodes.

Step 3: Generate the random weights and biases for each adding node and get the candidate basis function $\left\{ {g_{N + 1}^1,g_{N + 1}^2, \ldots ,g_{N + 1}^{{G_{\max }}}} \right\}$.

Step 4: Substitute $\left\{ {g_{N + 1}^1,g_{N + 1}^2, \ldots ,g_{N + 1}^{{G_{\max }}}} \right\}$ into (\ref{eq16}) and seek the nodes satisfying the inequality constraint.

Step 5: Define a set of variables $\xi _{j+1}^{*}\text{=}\left[ \xi _{j+1,1}^{*},...,\xi _{j+1,L}^{*} \right]$ to select the basis function making the training error converge as soon as possible, that is,
\begin{equation} \label{eq28}
\begin{array}{l}
\xi _{N + 1,q}^ *  = \frac{{{{\left( {{e_{N,q}},{g_{N + 1}}} \right)}^2}}}{{{{{{\left( {\left\| {{g_{N + 1}}} \right\|_{}^2{\rm{ + }}C} \right)}^2}} \mathord{\left/
 {\vphantom {{{{\left( {\left\| {{g_{N + 1}}} \right\|_{}^2{\rm{ + }}C} \right)}^2}} {\left( {\left\| {{g_{N + 1}}} \right\|_{}^2{\rm{ + 2}}C} \right)}}} \right.
 \kern-\nulldelimiterspace} {\left( {\left\| {{g_{N + 1}}} \right\|_{}^2{\rm{ + 2}}C} \right)}}}}\\
{\kern 1pt} {\kern 1pt} {\kern 1pt} {\kern 1pt} {\kern 1pt} {\kern 1pt} {\kern 1pt} {\kern 1pt} {\kern 1pt} {\kern 1pt} {\kern 1pt} {\kern 1pt} {\kern 1pt} {\kern 1pt} {\kern 1pt} {\kern 1pt} {\kern 1pt} {\kern 1pt} {\kern 1pt} {\kern 1pt} {\kern 1pt} {\kern 1pt} {\kern 1pt} {\kern 1pt} {\kern 1pt} {\kern 1pt} {\kern 1pt} {\kern 1pt} {\kern 1pt} {\kern 1pt} {\kern 1pt} {\kern 1pt} {\kern 1pt} {\kern 1pt}  - (1 - r - {\mu _{N + 1}})e_{N,q}^ \top e_{N,q}^{}.
\end{array}
\end{equation}

Step 6: Determine the adding node with the largest positive value of $\xi _{N+1}^{*}\text{=}\sum\limits_{q=1}^{L}{\xi _{N+1,q}^{*}}$ and construct the feedback matrix according to (\ref{eq5}).

Step 7: Calculate the output weights based on (\ref{eq13}) and (\ref{eq14}).

Step 8: Calculate the current residual error $e_{N+1}$ and renew ${{e}_{0}}:=e_{N+1}$, $N=N+1$. Repeat steps 3-7 until satisfying the terminal conditions.

A complete algorithm description of the RSC with hybrid regularization is given in Algorithm 1. 

\begin{remark}
The proposed method introduces a novel inequality constraint and utilizes the regularized least squares method to determine the output weights. The fundamental reservoir structure of the original RSCN is preserved so that the echo state property is naturally inherited. This property enables the network to maintain a stable dynamic response despite minor variations in the input, thereby enhancing the model's reliability.
\end{remark}
\section{Experimental results}
In this section, the effectiveness of our proposed method is verified on a nonlinear system identification task and two industrial data modelling problems. Performance comparisons are conducted among several models, including the original ESN, RSCN, and their variants with L1 and L2 regularization, termed ESN-L1,2, and RSCN-L1,2. Furthermore, we also consider variants processed with LASSO, denoted as LASSO-ESN, LASSO-RSCN, LASSO-ESN-L2, and LASSO-RSCN-L2. The training and testing performance of these models is assessed using the normalized root mean square error (NRMSE), which is formulated as follows:
\begin{equation} \label{eq29}
NRMSE=\sqrt{\frac{\sum\limits_{n=1}^{{{n}_{max}}}{{{\left( \mathbf{y}\left( n \right)-\mathbf{t}\left( n \right) \right)}^{2}}}}{{{n}_{max}}\operatorname{var}\left( \mathbf{t} \right)}},
\end{equation}where $\operatorname{var}\left( \mathbf{t} \right)$ denotes the variance of the desired output. In addition, the cross-validation is employed in the experiments to obtain the optimal ${C_L}$ in (\ref{eq121}). We calculated the sum of the absolute values of the variable feature coefficients corresponding to different ${C_L}$, that is,
\begin{equation} \label{eq291}
P = \sum\limits_{k = 1}^K {\sum\limits_{d = 1}^{d_B^K} {\left| {p_d^{d_B^k}} \right|} } ,
\end{equation}where $K$ is the number of input variables, ${d_B^K}$ is the corresponding order, ${p_d^{d_B^k}}$ denotes the feature coefficient associated with the $d$-th order of the $k$-th variable. Then, the regularization coefficient that maximizes $P$ is determined, and the corresponding order variables are selected.

\begin{table*}[htbp]
\caption{The range of network parameters for different modelling tasks.} \label{tb1}
\centering
\begin{tabular}{cccc}
\hline
Modelling tasks                   & The maximum reservoir size ${N_{\max }}$ & Reservoir sparsity & Scaling factor $\alpha $ \\ \hline
Nonlinear system identification & 300                          & 0.01-0.05          & 0.8-0.95       \\
Industry case 1                   & 200                          & 0.03-0.09          & 0.5-0.95       \\
Industry case 2                   & 300                          & 0.01-0.05          & 0.6-0.95       \\ \hline
\end{tabular}
\end{table*}

To determine certain hyperparameters, a series of preliminary experiments was conducted to establish the corresponding value ranges. Following this, the grid search method was used to select the optimal network parameters. By systematically exploring the predefined hyperparameter space, we can identify the parameter combinations that optimize model performance. The parameter ranges for various modelling tasks are presented in Table~\ref{tb1}. Each experiment consisted of 50 independent trials conducted under identical conditions to ensure the repeatability and reliability of the results. Moreover, we recorded the training and testing NRMSE, as well as the training time. Model performance is evaluated based on the mean and variance of these metrics.

Specifically, for the RSC frameworks, the following parameters are utilized: the weight scale sequence is defined as $\left\{ 0.1,0.5,1,5,10,30,50,100 \right\}$, the contractive sequence is given by $r=\left[ 0.9,0.99,0.999,0.9999,0.99999 \right]$, the maximum number of stochastic configurations is set to ${{G}_{\max }}=100$, the training tolerance is specified as  ${{\varepsilon }}={{10}^{-5}}$, and the initial reservoir size is set to 5. 

\subsection{Nonlinear system identification}
In this simulation, the nonlinear dynamic system is expressed as
\begin{equation} \label{eq33}
\begin{array}{l}
y\left( {n + 1} \right) = 
\frac{{y\left( n \right)y\left( {n - 1} \right)y\left( {n - 2} \right)u\left( {n - 1} \right)\left( {y\left( {n - 2} \right) - 1 + u\left( n \right)} \right)}}{{1 + y\left( {n - 1} \right)y\left( {n - 1} \right) + y\left( {n - 2} \right)y\left( {n - 2} \right)}}.
\end{array}
\end{equation}
The input $u\left( n \right)$ is generated from the uniform distribution $\left[ { - 0.7,0.7} \right]$ in the training phase, and the initial output values are set to $y\left( 1 \right)=y\left( 2 \right)=y\left( 3 \right)=0,$ $y\left( 4 \right)=0.1.$ In the testing phase, the input is generated by 
\begin{equation} \label{eq34}
u\left( n \right) = \left\{ {\begin{array}{*{20}{c}}
{0.3*\sin \left( {\frac{{\pi n}}{{125}}} \right) + 0.2*\sin \left( {\frac{{\pi n}}{{25}}} \right){\kern 1pt} {\kern 1pt} {\kern 1pt} {\kern 1pt} {\kern 1pt} {\kern 1pt} {\kern 1pt} {\kern 1pt} 0 \le n < 500}\\
{0.6*\sin \left( {\frac{{\pi n}}{{25}}} \right){\kern 1pt} {\kern 1pt} {\kern 1pt} {\kern 1pt} {\kern 1pt} {\kern 1pt} {\kern 1pt} {\kern 1pt} {\kern 1pt} {\kern 1pt} {\kern 1pt} {\kern 1pt} {\kern 1pt} {\kern 1pt} {\kern 1pt} {\kern 1pt} {\kern 1pt} {\kern 1pt} {\kern 1pt} {\kern 1pt} {\kern 1pt} {\kern 1pt} {\kern 1pt} {\kern 1pt} {\kern 1pt} {\kern 1pt} {\kern 1pt} {\kern 1pt} {\kern 1pt} {\kern 1pt} {\kern 1pt} {\kern 1pt} {\kern 1pt} {\kern 1pt} {\kern 1pt} {\kern 1pt} {\kern 1pt} {\kern 1pt} {\kern 1pt} {\kern 1pt} {\kern 1pt} {\kern 1pt} {\kern 1pt} {\kern 1pt} {\kern 1pt} {\kern 1pt} {\kern 1pt} {\kern 1pt} {\kern 1pt} {\kern 1pt} {\kern 1pt} {\kern 1pt} {\kern 1pt} {\kern 1pt} {\kern 1pt} {\kern 1pt} {\kern 1pt} {\kern 1pt} {\kern 1pt} {\kern 1pt} {\kern 1pt} {\kern 1pt} {\kern 1pt} {\kern 1pt} {\kern 1pt} {\kern 1pt} {\kern 1pt} {\kern 1pt} {\kern 1pt} {\kern 1pt} {\kern 1pt} {\kern 1pt} {\kern 1pt} {\kern 1pt} {\kern 1pt} 500 \le n \le 800}
\end{array}} \right.,
\end{equation}
and the initial output values are $y\left( 1 \right)=-0.3, y\left( 2 \right)=-0.1,$ $y\left( 3 \right)=0.3,$ $y\left( 4 \right)=0.$ A total of 2000, 1000, and 800 samples are utilized for training, validation, and testing, respectively. The first 10 samples of each set are washed out.
\setlength{\abovecaptionskip}{-0.3cm}
\begin{figure}[ht] 
	\centering
	\includegraphics[angle=90,width=9.5cm]{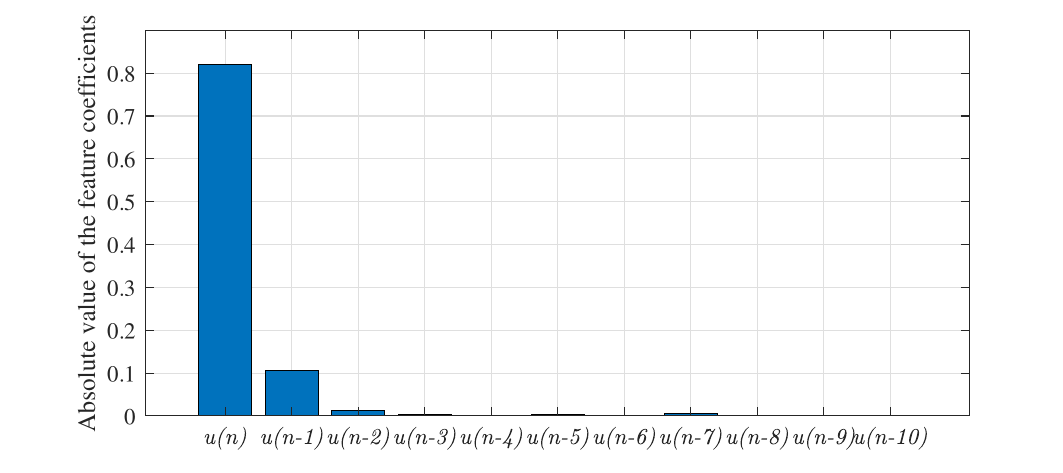}
	\caption{The distribution of feature coefficients corresponding to each order variable for the nonlinear system identification task.}
	\label{fig2}
\end{figure}
\begin{figure}[ht]
\vspace{-0.3cm}
	\centering
	\subfloat[Prediction results]{\includegraphics[width=9cm]{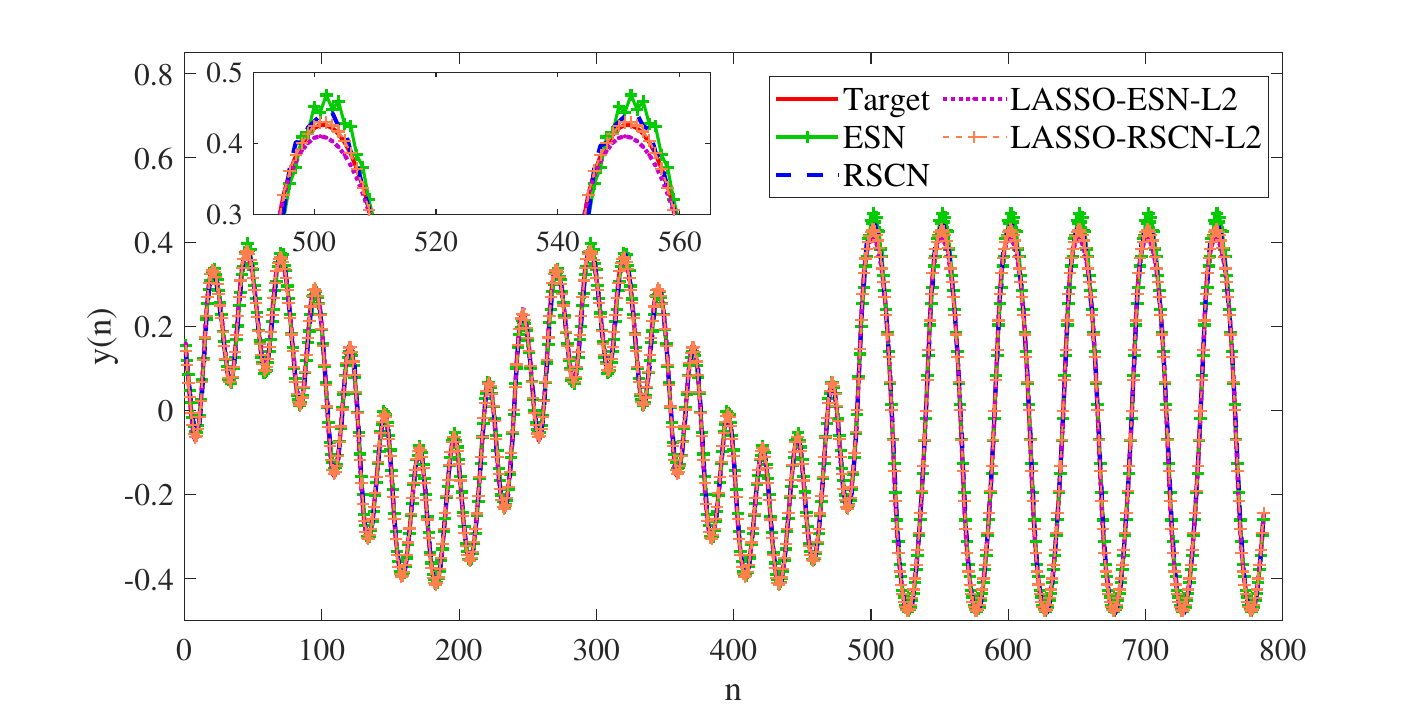}}\\ \vspace{-0.2cm}
	\subfloat[Prediction errors]{\includegraphics[width=9cm]{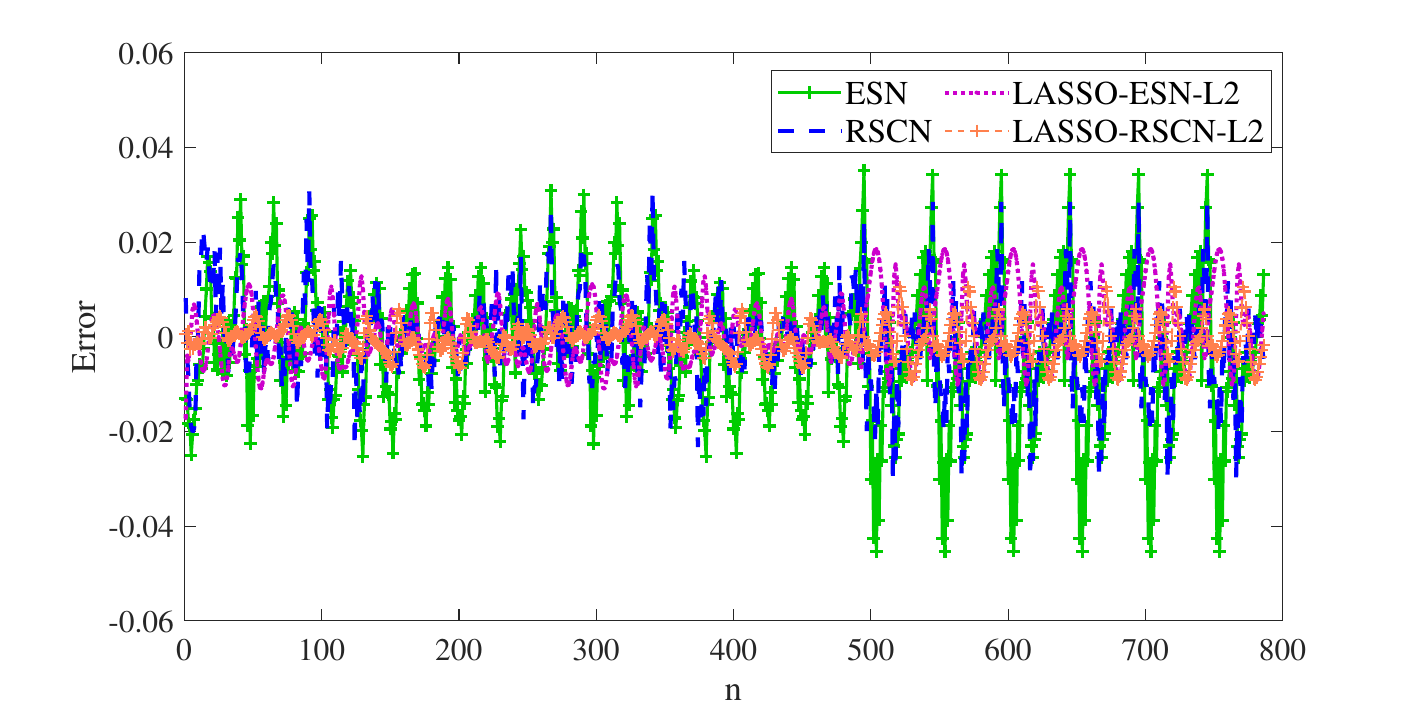}}
	\caption{The prediction fitting curves and error values of different models for the nonlinear system identification task.}
	\label{fig3}
    \vspace{-0.2cm}
\end{figure}

For the original ESN, RSCN and their regularized variants, the plant output $y\left( n\text{+}1 \right)$ is predicted by $\left[ y\left( n \right),u\left( n \right) \right]$. For the LASSO-based frameworks, it is crucial to first extract important order information. As illustrated in Fig.~\ref{fig2}, we establish ten-order delays and compute the feature coefficients. The results clearly indicate that the feature coefficient of the order variable $u\left( n \right)$ is significantly higher than those of the other input orders. This suggests that the current input exerts a more substantial influence on the system output, playing a dominant role in output prediction. Further analysis of these feature coefficients will enhance our understanding of the specific contributions of different order inputs to the system's behavior, thereby providing valuable insights for model selection and optimization.

\setlength{\abovecaptionskip}{5pt}
\begin{table*}[ht]
\caption{Performance comparison of different models on two nonlinear system identification tasks.} \label{tb2}
\centering
\begin{tabular}{ccccc}
\hline
Models        & Reservoir size & Training time   & Training NRMSE  & Testing NRMSE   \\ \hline
ESN           & 278            & \textbf{0.63861±0.03295} & 0.03183±0.00225 & 0.04859±0.00679 \\
ESN-L1,2      & 227            & 0.66098±0.08982 & 0.03756±0.00563 & 0.04207±0.00327 \\
LASSO-ESN     & 136            & 0.78039±0.07315 & 0.03052±0.00271 & 0.03953±0.00362 \\
LASSO-ESN-L2  & 119            & 0.74682±0.08283 & 0.03135±0.00308 & 0.03908±0.00573 \\
RSCN          & 148            & 2.98767±0.32301 & 0.02553±0.00593 & 0.03840±0.00236 \\
RSCN-L1,2     & 113            & 3.32891±0.58170 & 0.02629±0.00305 & 0.03376±0.00296 \\
LASSO-RSCN    & 93             & 3.34637±0.65106 & \textbf{0.01808±0.00297} & 0.03159±0.00229 \\
LASSO-RSCN-L2 & 87             & 3.28737±0.52367 & 0.01972±0.00198 & \textbf{0.02345±0.00116} \\ \hline
\end{tabular}
\vspace{-0.3cm}
\end{table*}

The prediction fitting curves and error values generated by various models for the nonlinear system identification tasks are shown in Fig.~\ref{fig3}. It is obvious that the LASSO-RSCN-L2 outperforms other methods in fitting the desired outputs, with the smaller error range. These findings highlight the effectiveness of our proposed approach in modelling nonlinear dynamic systems.

The detailed performance comparisons of the nonlinear system identification problem are presented in Table~\ref{tb2}. The experimental results demonstrate that LASSO regularization significantly reduces the reservoir size of the models while enhancing their overall performance, particularly as LASSO-RSCN exhibits the lowest training NRMSE. Notably, the proposed LASSO-RSCN-L2 model achieves outstanding performance, achieving the lowest testing NRMSE while maintaining model compactness. This hybrid regularized strategy preserves critical order variable information and reduces parameter fluctuations, increasing the model's reliability across different environments and conditions. Therefore, our proposed approach is well-suited for identifying nonlinear systems, especially those with dynamic unknown orders and ill-posed problems.
\begin{figure}[ht]
\vspace{-0.3cm}
        \centering
	\subfloat{\includegraphics[width=8cm]{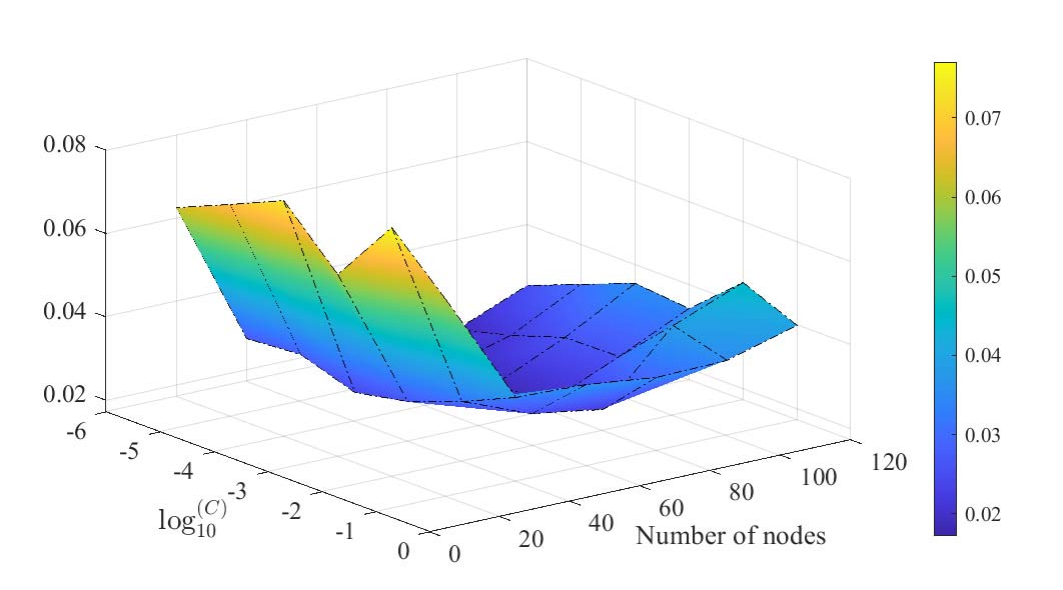}}
	\caption{The testing NRMSE surface map of LASSO-RSCN-L2 with different regularization coefficient $C$ and reservoir size on the nonlinear system identification task.}
	\label{fig4}
    \vspace{-0.2cm}
\end{figure}

The testing performance of the proposed LASSO-RSCN-L2 with different combinations of regularization coefficient $C$ and reservoir size $N$ is illustrated in Fig.~\ref{fig4}. The testing NRMSE varies with regularization coefficients, and the optimal results can be achieved when $N=80$, $C=0.001$ for the nonlinear system identification task. This finding not only demonstrates the model's sensitivity to parameter selection but also highlights the significance of appropriate regularization strategies in enhancing model performance. By carefully adjusting the reservoir size and regularization coefficient, the built model can effectively balance complexity and learning ability, resulting in improved outcomes in nonlinear dynamic modelling.

\subsection{Soft sensing of the butane concentration in the debutanizer column process}
\begin{figure}[htbp]
	\begin{center}
		\includegraphics[width=6cm]{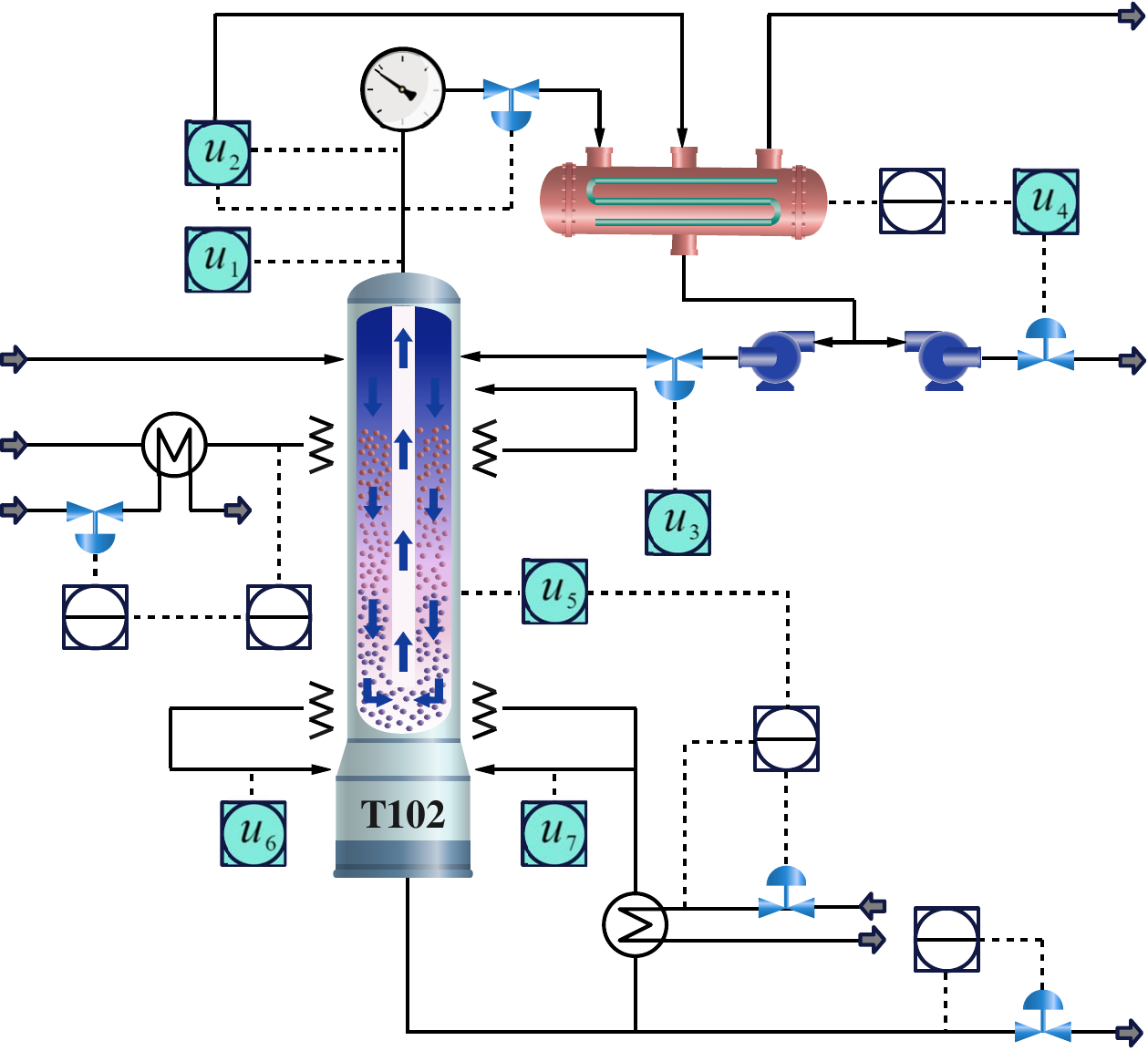}
		\caption{Flowchart of debutanizer column process.}
		\label{fig5}
	\end{center}
\end{figure}

The debutanizer column is an important equipment in the naphtha fractionation process, the main role is to separate butane (C4) from naphtha, the specific operation flow is shown in Fig.~\ref{fig5}. In industrial processes, butane concentration needs to be measured in real-time to minimize the butane content at the bottom of the tower. However, butane concentration is often difficult to detect directly and incurs a certain delay. It is necessary to establish an accurate soft sensing model. Table~\ref{tb3} displays the relevant auxiliary variables in the production process. In \cite{ref29}, Fortuna et al. presented a well-designed combination of variables to obtain the butane concentration $y\left( n \right)$, that is,
\vspace{-0.3cm}

\begin{small}
    \begin{equation} \label{eq35}
\begin{array}{l}
y\left( n \right) = f\left( {{u_1}\left( n \right),} \right.{u_2}\left( n \right),{u_3}\left( n \right),{u_4}\left( n \right),{u_5}\left( n \right),{u_5}\left( {n - 1} \right),\\
{\kern 1pt} {\kern 1pt} {\kern 1pt} {\kern 1pt} {\kern 1pt} {\kern 1pt} {\kern 1pt} {\kern 1pt} {\kern 1pt} {\kern 1pt} {\kern 1pt} {\kern 1pt} {\kern 1pt} {\kern 1pt} {\kern 1pt} {\kern 1pt} {\kern 1pt} {\kern 1pt} {\kern 1pt} {\kern 1pt} {\kern 1pt} {\kern 1pt} {\kern 1pt} {\kern 1pt} {\kern 1pt} {\kern 1pt} {\kern 1pt} {\kern 1pt} {\kern 1pt} {\kern 1pt} {\kern 1pt} {\kern 1pt} {\kern 1pt} {\kern 1pt} {\kern 1pt} {\kern 1pt} {\kern 1pt} {\kern 1pt} {\kern 1pt} {\kern 1pt} {\kern 1pt} {\kern 1pt} {\kern 1pt} {\kern 1pt} {\kern 1pt} {\kern 1pt} {\kern 1pt} {\kern 1pt} {\kern 1pt} {\kern 1pt} {u_5}\left( {n - 2} \right),{u_5}\left( {n - 3} \right),\left( {{u_1}\left( n \right) + {u_2}\left( n \right)} \right)/2,\\
{\kern 1pt} {\kern 1pt} {\kern 1pt} {\kern 1pt} {\kern 1pt} {\kern 1pt} {\kern 1pt} {\kern 1pt} {\kern 1pt} {\kern 1pt} {\kern 1pt} {\kern 1pt} {\kern 1pt} {\kern 1pt} {\kern 1pt} {\kern 1pt} {\kern 1pt} {\kern 1pt} {\kern 1pt} {\kern 1pt} {\kern 1pt} {\kern 1pt} {\kern 1pt} {\kern 1pt} {\kern 1pt} {\kern 1pt} {\kern 1pt} {\kern 1pt} {\kern 1pt} {\kern 1pt} {\kern 1pt} {\kern 1pt} {\kern 1pt} {\kern 1pt} {\kern 1pt} {\kern 1pt} {\kern 1pt} {\kern 1pt} {\kern 1pt} {\kern 1pt} {\kern 1pt} {\kern 1pt} {\kern 1pt} {\kern 1pt} {\kern 1pt} {\kern 1pt} {\kern 1pt} \left. {{\kern 1pt} {\kern 1pt} {\kern 1pt} y\left( {n - 1} \right),y\left( {n - 2} \right),y\left( {n - 3} \right),y\left( {n - 4} \right)} \right).
\end{array}
\end{equation}
\end{small}A total of 2394 samples are generated, where 1500 samples with $n\in \left[ 1,1500 \right]$ are chosen as the training set, 894 samples with $n\in \left[ 1501,2394 \right]$ are chosen as the testing set, and Gaussian white noise is added to the testing set to generate the validation set. The first 100 samples of each set are washed out.
\vspace{-0.5cm}
\begin{table}[ht]
\vspace{-0.3cm}
 \centering
  \caption{Process auxiliary variables for the debutanizer column.}
  \label{tb3}
\begin{tabular}{cc}
\hline
Input variables & Variable description      \\ \hline
${{u_1}}$             & tower top   temperature   \\
${{u_2}}$              & tower top   pressure      \\
${{u_3}}$             & tower top   reflux flow   \\
${{u_4}}$             & tower top product outflow \\
${{u_5}}$             & 6-th tray   temperature   \\
${{u_6}}$            & tower bottom temperature  \\
${{u_7}}$             & tower bottom pressure     \\ \hline
\end{tabular}
\end{table}
\begin{figure}[ht] 
	\centering
	\includegraphics[width=9cm]{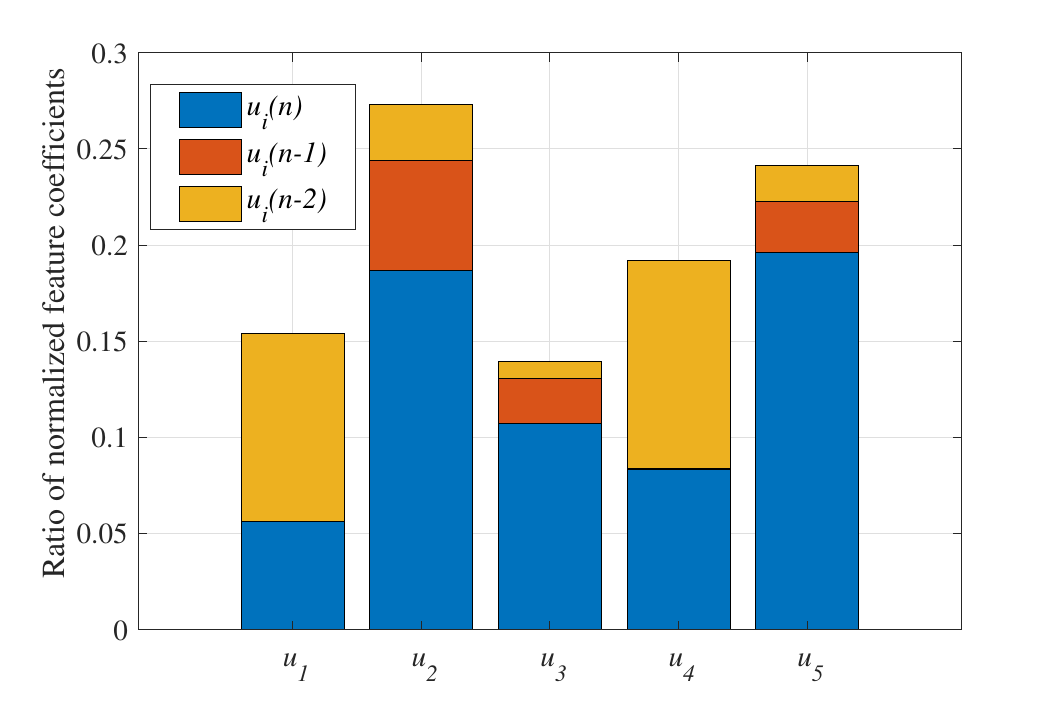}
	\caption{The distribution of feature coefficients corresponding to each order variable for the debutanizer column process.}
	\label{fig6}
    \vspace{-0.5cm}
\end{figure}

For the non-LASSO-based model, $\left[ {{u}_{1}}\left( n \right), \right.{{u}_{2}}\left( n \right),{{u}_{3}}\left( n \right),$ $\left. {{u}_{4}}\left( n \right),{{u}_{5}}\left( n \right),y\left( n-1 \right) \right]$ is used to predict $y\left( n \right)$. Then, we have implemented second-order delays for each input feature to better capture the dynamic characteristics of the time series and to conduct a comprehensive analysis of each order variable's contribution to the model output. As shown in Fig.~\ref{fig6}, by calculating the corresponding feature coefficients, the most suitable inputs are determined to be $\left[ {{u_1}\left( {n - 2} \right),{u_2}\left( n \right),{u_3}\left( n \right),{u_4}\left( {n - 2} \right),{u_5}\left( n \right)} \right]$. This analysis provides crucial insights that inform model selection and optimization, enabling us to enhance the predictive performance and overall effectiveness of the model in capturing the underlying system dynamics. These insights facilitate more informed decisions regarding order identification and input adjustments, ultimately leading to improved accuracy and reliability in predictions.
\begin{figure}[ht] 
\vspace{-0.3cm}
	\centering
	\subfloat[Prediction results]{\includegraphics[width=9cm]{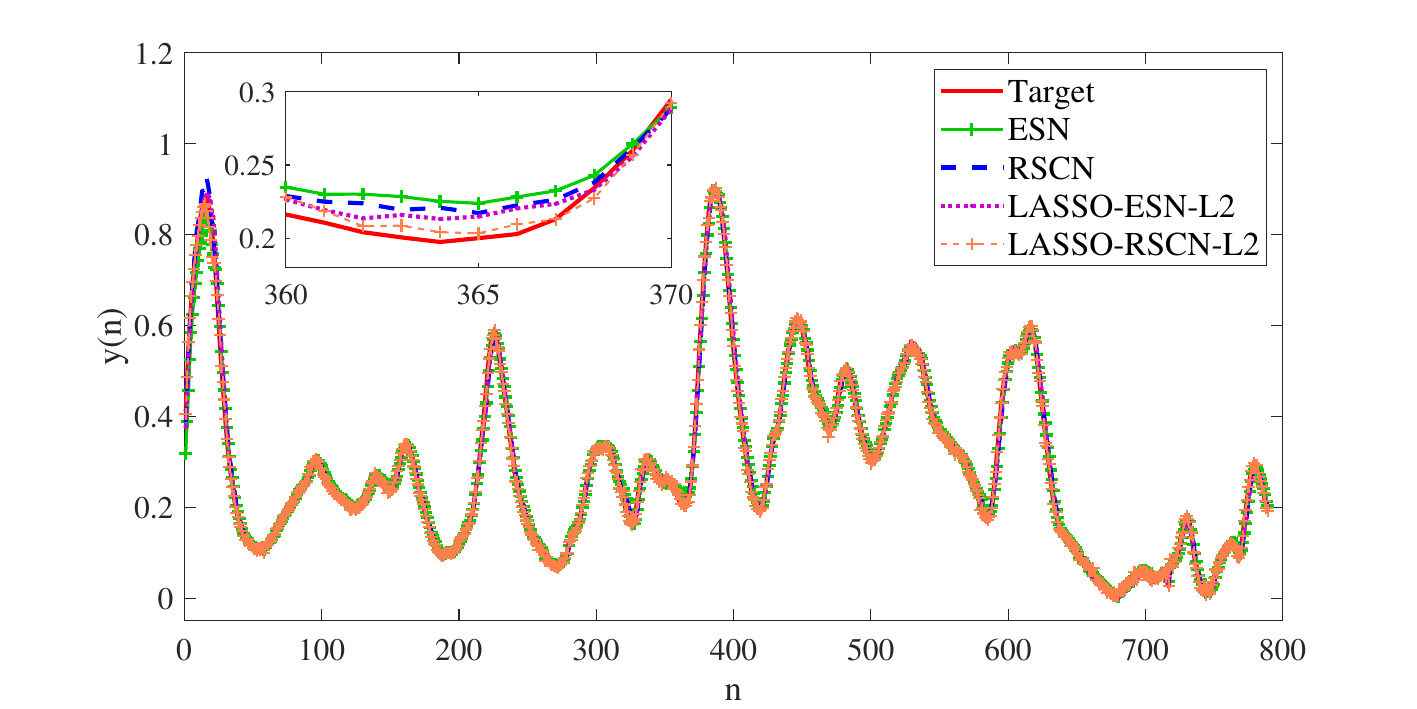}}\\ \vspace{-0.2cm}
	\subfloat[Prediction errors]{\includegraphics[width=9cm]{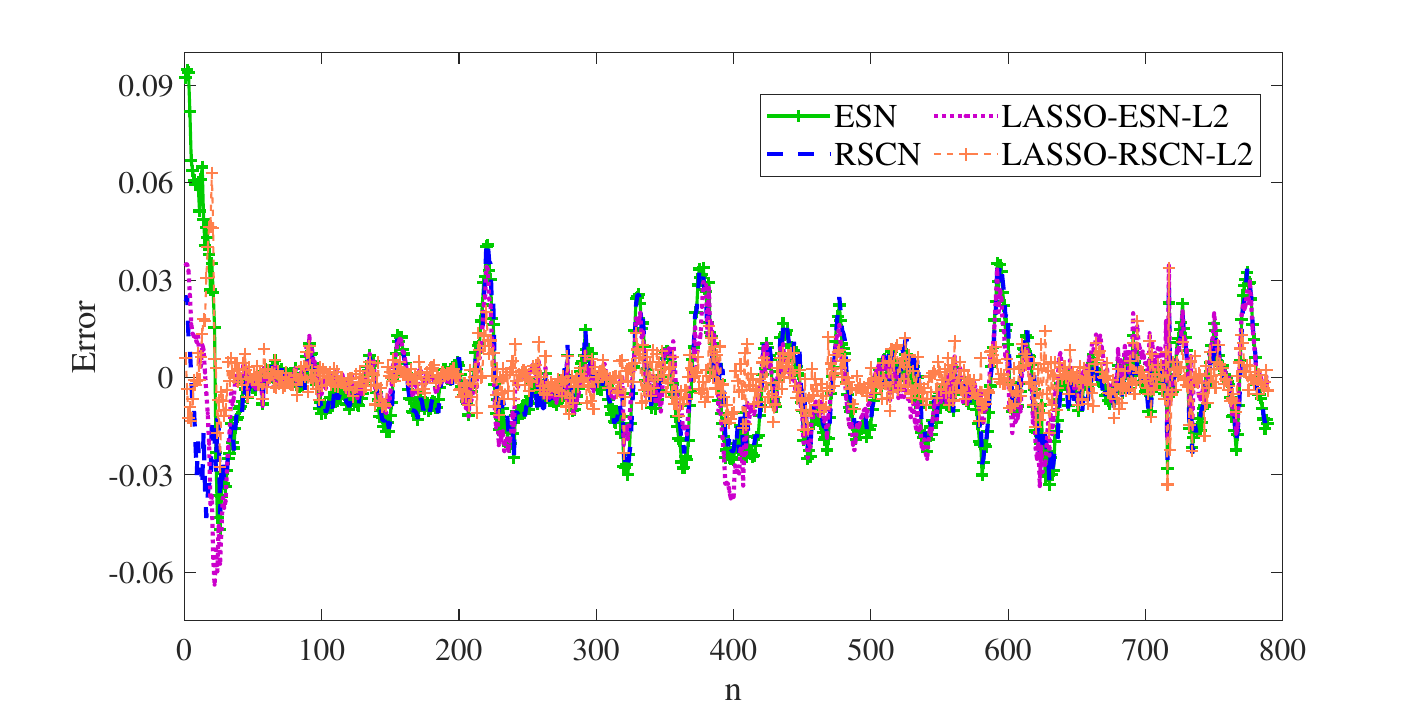}}
	\caption{The prediction fitting curves and error values of different models for the debutanizer column process.}
	\label{fig7}
\end{figure}

Fig.~\ref{fig7} depicts the prediction curves and errors of different models for the soft sensing of the debutanizer column process. It is evident that our proposed LASSO-RSCN-L2 model exhibits superior prediction accuracy compared to other methods, with a notably smaller error range. This enhanced performance facilitates reliable real-time predictions in practical applications, which is essential for monitoring and adjusting production processes. Such capabilities underscore the effectiveness of the proposed method in industrial nonlinear dynamic modelling.

\subsection{Short-term power load forecasting}
\begin{figure}[ht]
	\begin{center}
		\includegraphics[width=8cm]{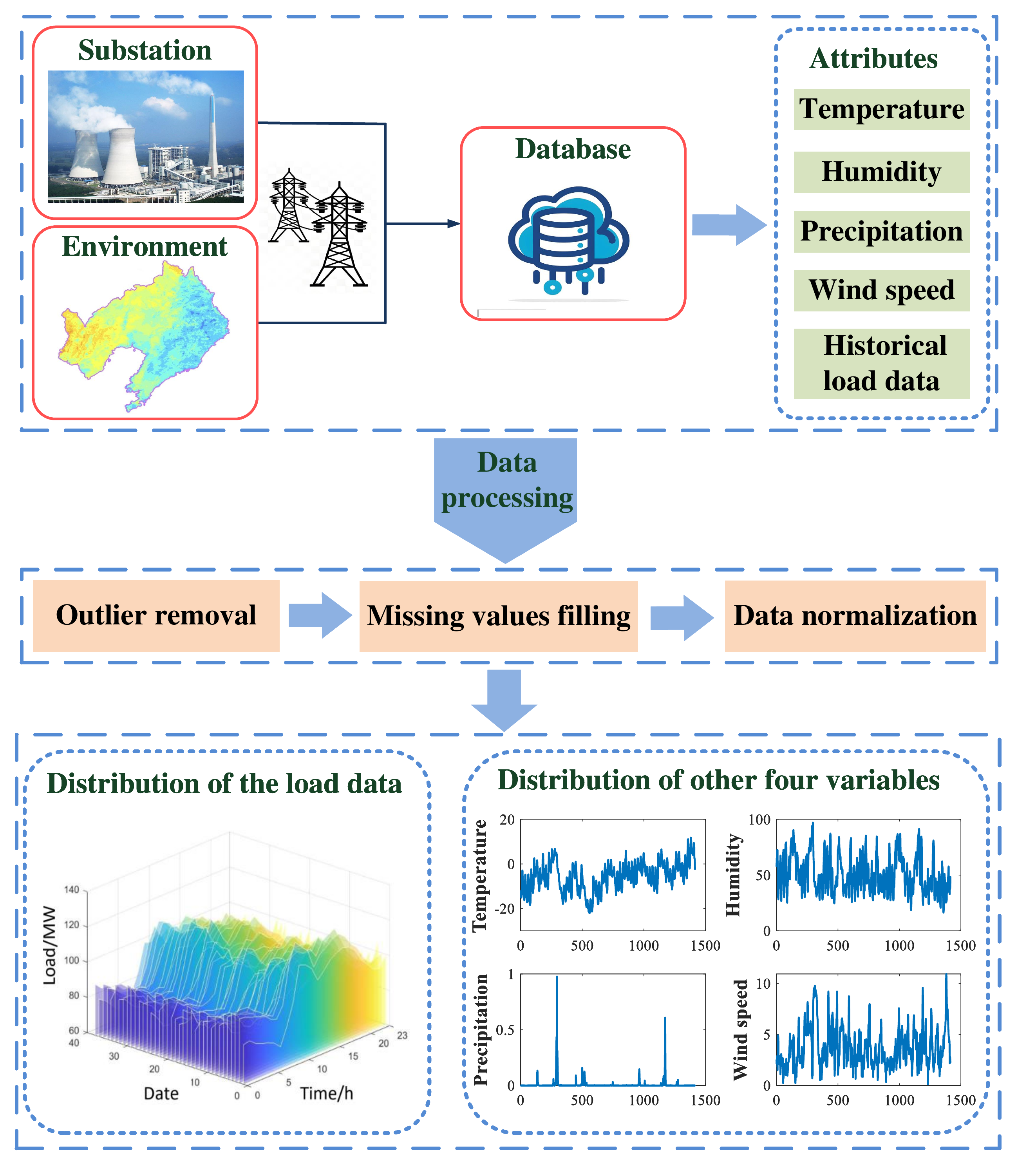}
		\caption{Flowchart of the data collection and processing for the short-term
power load forecast.}
		\label{fig8}
	\end{center}
    \vspace{-0.5cm}
\end{figure}
Short-term power load forecasting is essential for decision-makers to efficiently allocate power generation, transmission, and distribution resources to meet varying power demands and enhance energy efficiency. Environmental variables play a crucial role in electricity consumption behaviour and power system operations. By analyzing these factors alongside historical load data, predictive models can be developed for future load forecasting. This study utilizes the load data from a 500kV substation in Liaoning Province, China, over 59 days from January to February 2023. Environmental variables such as temperature ${u_1}$, humidity ${u_2}$, precipitation ${u_3}$, and wind speed ${u_4}$ are taken into account to predict the power load $y$. The process of data collection and analysis is depicted in Fig.~\ref{fig8}. This dataset consists of 1415 samples, with 1000 for training and 415 for testing. Gaussian noise is added to the testing set to create a validation set. The first 30 samples from each set are washed out.

\begin{figure}[ht] 
\vspace{-0.3cm}
	\centering
	\includegraphics[width=9cm]{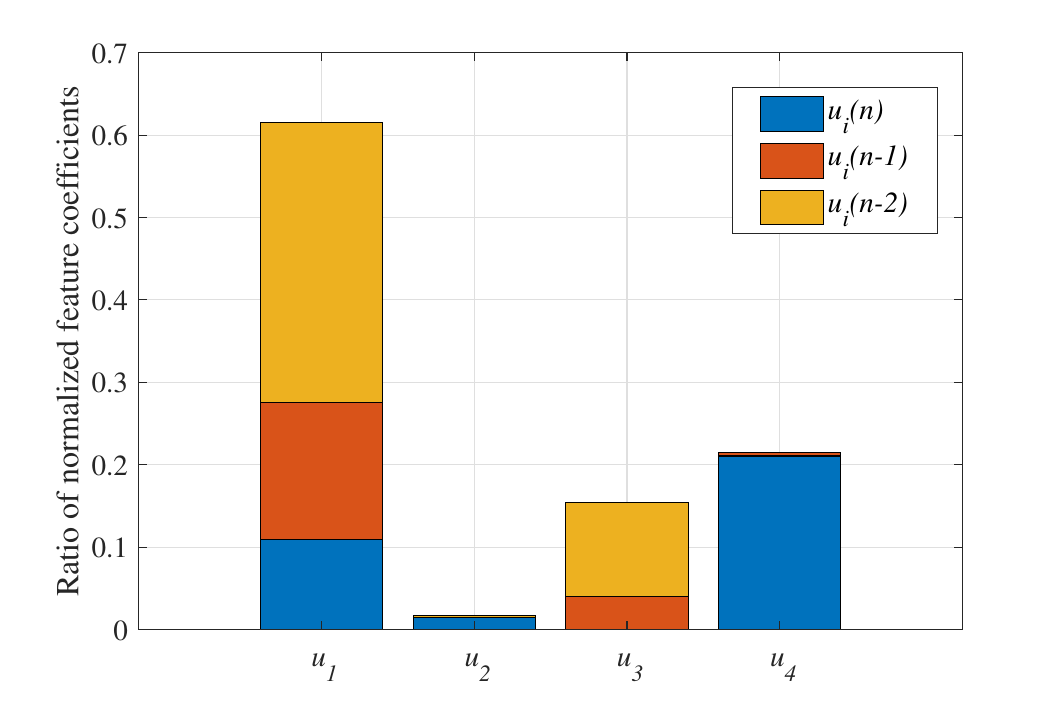}
	\caption{The distribution of feature coefficients corresponding to each order variable for the power load forecasting.}
	\label{fig9}
    \vspace{-0.5cm}
\end{figure}

ESN, RSCN, and their variants utilize $\left[ {{u}_{1}}\left( n \right), \right.{{u}_{2}}\left( n \right),$ $\left. {{u}_{3}}\left( n \right), {{u}_{4}}\left( n \right), y\left( n-1 \right) \right]$ to predict $y\left( n \right)$. For LASSO-based frameworks, we first need to perform order identification. Fig.~\ref{fig9} presents the distribution of feature coefficients for different order variables, revealing that $\left[ {{u_1}\left( {n - 2} \right),{u_2}\left( n \right),{u_3}\left( {n - 2} \right),{u_4}\left( n \right)} \right]$ plays a key role in the prediction. This finding deepens our understanding of the system's dynamic behavior and the intricate relationships among inputs. It not only facilitates the optimization of the model structure but also significantly enhances the model's responsiveness and stability. Consequently, the model offers more reliable decision support in the face of order uncertainty and dynamic changes.

\begin{figure}[ht] 
	\centering
	\subfloat[Prediction results]{\includegraphics[width=9cm]{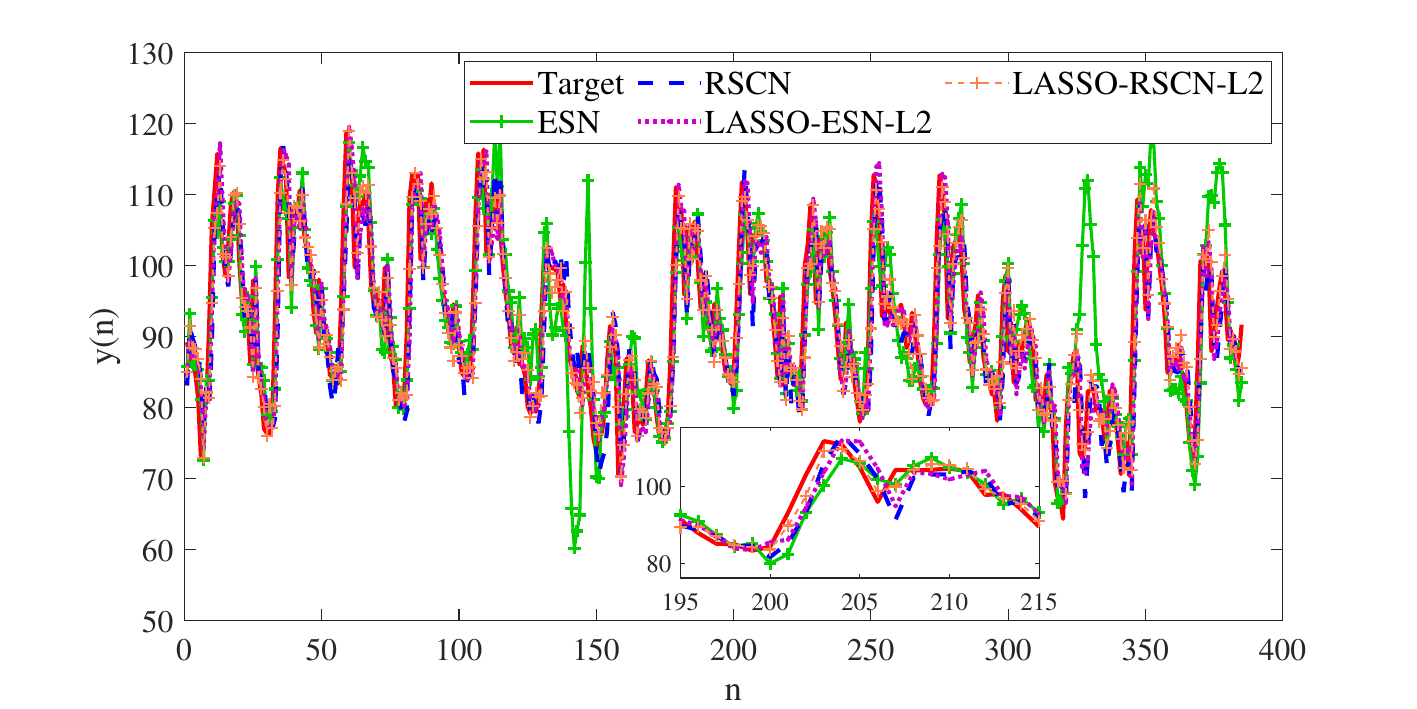}}\\ \vspace{-0.2cm}
	\subfloat[Prediction errors]{\includegraphics[width=9cm]{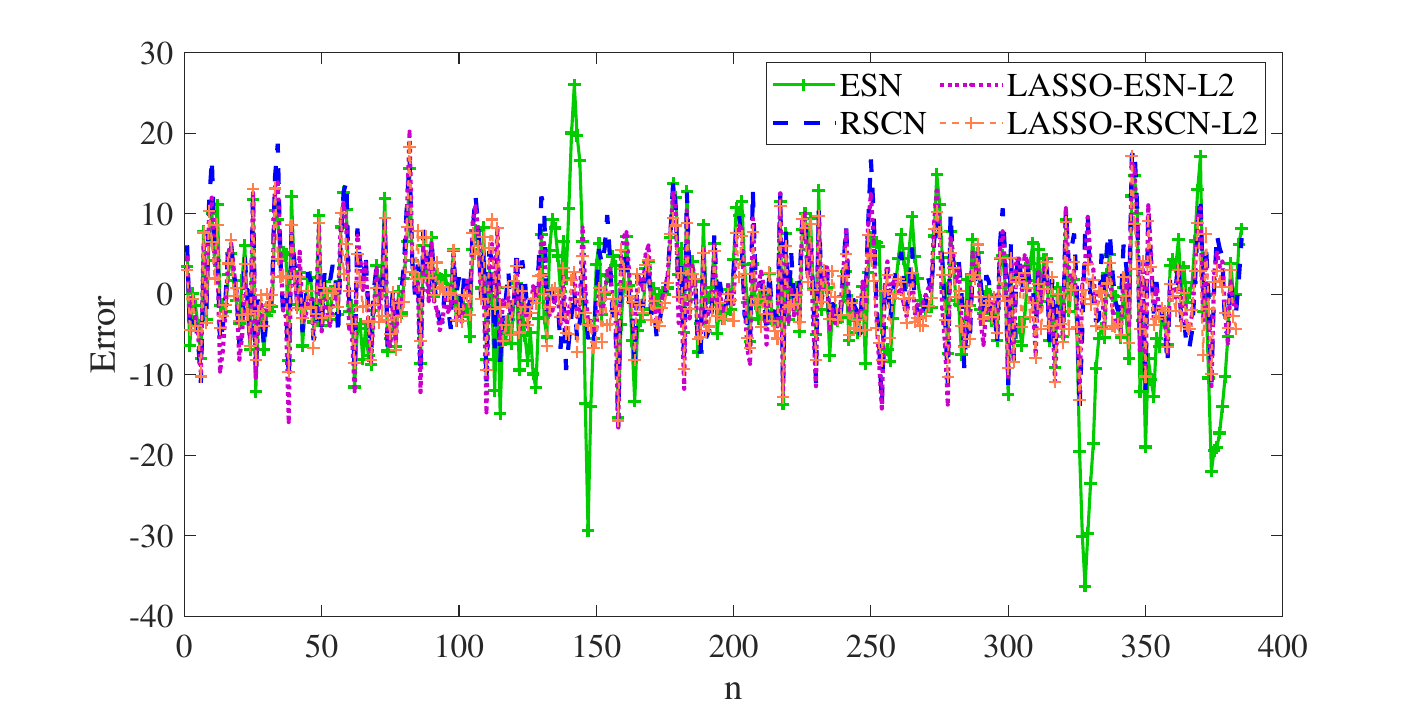}}
	\caption{The prediction fitting curves and error values of different models for the power load forecasting.}
	\label{fig10}
\end{figure}

Fig.~\ref{fig10} plots the prediction results and errors of various models on the short-term power load forecasting. It can be seen that the outputs generated by LASSO-RSCN-L2 align more closely with the target outputs and demonstrate smaller prediction errors, ranging from $\left[ -16.53,12.12 \right]$. These results highlight the efficacy of the proposed methods in capturing complex dynamics within the industrial processes.
\begin{table*}[ht]
\caption{Performance comparison of different models on the two industry cases.} \label{tb4}
\centering
\begin{tabular}{cccccc}
\hline
Datasets                & Models        & Reservoir size & Training time   & Training NRMSE  & Testing NRMSE   \\ \hline
\multirow{8}{*}{Case 1} & ESN           & 213            & 0.40819±0.12035 & 0.04085±0.00137 & 0.08427±0.01144 \\
                        & ESN-L1,2      & 157            & 0.43006±0.10432 & 0.03768±0.00161 & 0.07362±0.00879 \\
                        & LASSO-ESN     & 125            & 0.32829±0.11235 & 0.03203±0.00129 & 0.06827±0.00537 \\
                        & LASSO-ESN-L2  & 103            & \textbf{0.32281±0.13198} & 0.03385±0.00102 & 0.06811±0.00763 \\
                        & RSCN          & 87             & 2.23734±0.20122 & 0.02734±0.00100 & 0.06793±0.00411 \\
                        & RSCN-L1,2     & 94             & 2.30281±0.25219 & 0.02853±0.00167 & 0.06265±0.00376 \\
                        & LASSO-RSCN    & 56             & 1.88296±0.21003 & \textbf{0.02391±0.00029} & 0.04305±0.00398 \\
                        & LASSO-RSCN-L2 & 28             & 1.60479±0.18993 & 0.02698±0.00086 & \textbf{0.03926±0.00383} \\ \hline
\multirow{8}{*}{Case 2} & ESN           & 103            & 0.11881±0.01548 & 0.24338±0.02189 & 0.64522±0.07172 \\
                        & ESN-L1,2      & 83             & 0.09312±0.00937 & 0.21096±0.02005 & 0.57328±0.05631 \\
                        & LASSO-ESN     & 67             & \textbf{0.09182±0.00718} & 0.20029±0.01876 & 0.53623±0.04938 \\
                        & LASSO-ESN-L2  & 61             & 0.09286±0.01137 & 0.21672±0.02531 & 0.52151±0.06733 \\
                        & RSCN          & 65             & 0.68257±0.03141 & \textbf{0.19254±0.00867} & 0.51078±0.03685 \\
                        & RSCN-L1,2     & 68             & 0.74521±0.01829 & 0.20503±0.00335 & 0.50462±0.02161 \\
                        & LASSO-RSCN    & 57             & 0.69876±0.02003 & 0.20427±0.00136 & 0.46445±0.00287 \\
                        & LASSO-RSCN-L2 & 36             & 0.66325±0.01673 & 0.20265±0.00217 & \textbf{0.44771±0.00203} \\ \hline
\end{tabular}
\vspace{-0.3cm}
\end{table*}

\begin{figure}[ht]
        \centering
        \subfloat[Case 1]{\includegraphics[width=8cm]{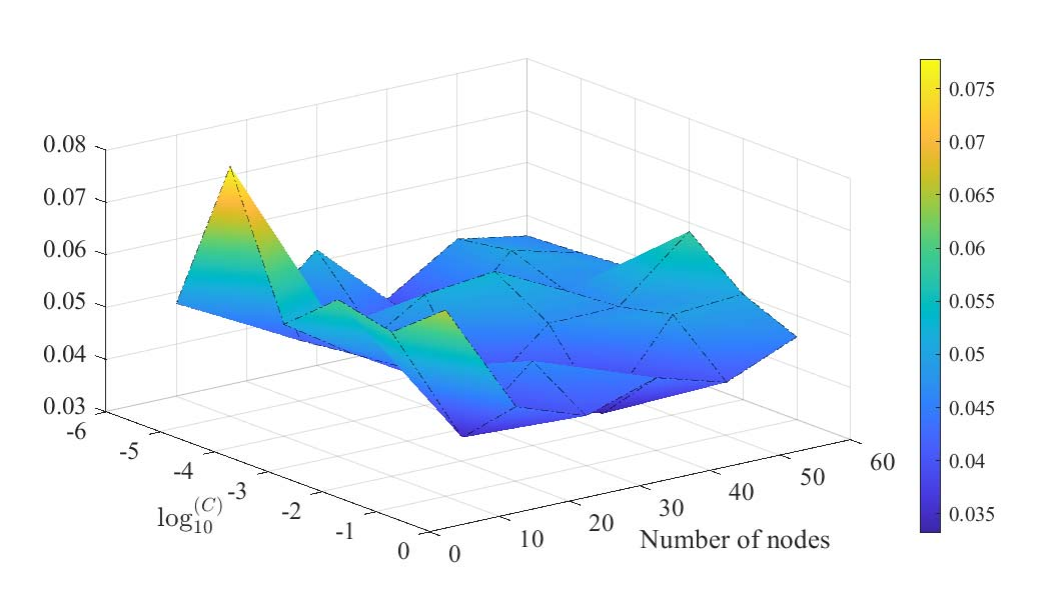}}\\
	\subfloat[Case 2]{\includegraphics[width=8cm]{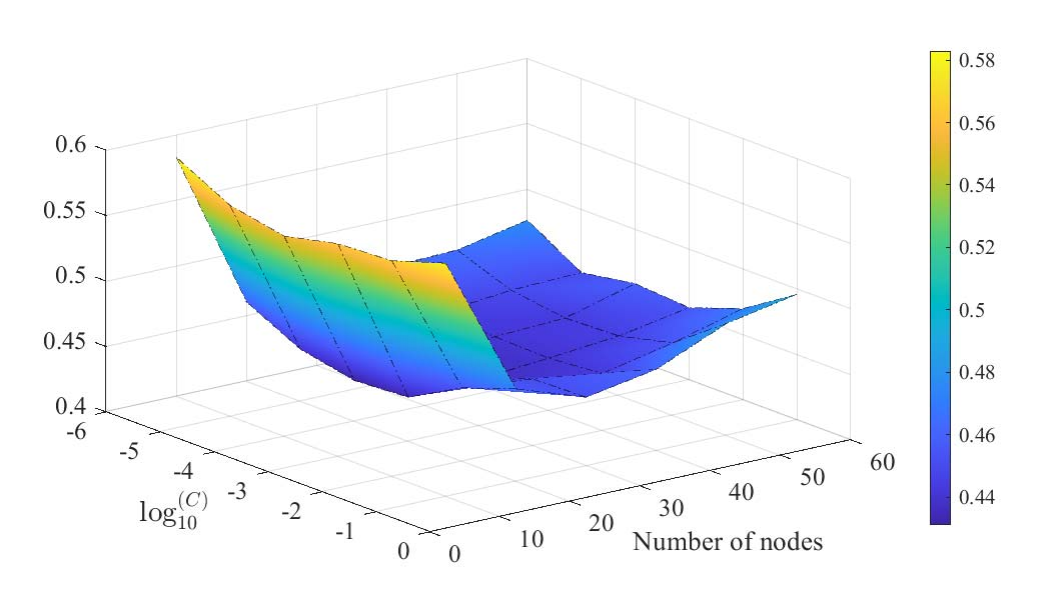}}
	\caption{The testing NRMSE surface map of the LASSO-RSCN-L2 with different regularization coefficient $C$ and reservoir size $N$ on the two industry cases.}
	\label{fig11}
    \vspace{-0.5cm}
\end{figure}

To facilitate a comparative analysis of the hybrid regularized RSCN with other models, we summarize the experimental results in Table~\ref{tb4}. The LASSO-RSCN-L2 achieves the best testing performance across the two industrial modelling tasks. Furthermore, the introduction of LASSO regularization contributes to more compact reservoirs. These findings highlight the promising potential of the proposed approaches in managing complex dynamics within practical industrial applications. Real industrial processes are often intricate and varied, which can lead to ill-posed problems. Regularization techniques help mitigate this issue by constraining the model's complexity through a penalty term, thereby enhancing its ability to generalize to unknown data streams. This ultimately enhances the efficiency and reliability of industrial systems for operation and control.

The testing performance of the proposed method with different combinations of regularization coefficient $C$ and reservoir size $N$ is shown in Fig.~\ref{fig11}. Notably, the testing NRMSE is sensitive to both parameters, with optimal prediction results observed at $N=30$, $C=0.01$ and $N=30$, $C=0.001$ for the two industry tasks, respectively. By conducting a comprehensive examination of the effects of various parameter settings, we can gain a deeper insight into the model's flexibility and boost its ability to generalize in real-world situations. This contributes to more dependable strategic guidance for industrial applications, ensuring that systems can quickly respond to the complexities and dynamics of real-world scenarios.

\subsection{Discussions}
The proposed LASSO-RSCN-L2 can be viewed as a hybrid framework consisting of two main construction phases: the first phase involves order variable extraction and linear regression modelling based on the LASSO, reducing the model complexity and enhancing interpretability. The second phase utilizes an RSCN with L2 regularization to approximate the model's output residuals, further improving predictive accuracy. Distinguished with the RSCN-L1,2, which integrates L1 and L2 penalty terms into the loss function, that is,
\begin{equation}
\mathop {\min }\limits_{{\bf{W}}_{{\mathop{\rm out}\nolimits} }^{}} \left\| {{\bf{W}}_{{\mathop{\rm out}\nolimits} }^{}{{{\bf{\hat X}}}_{N + 1}} - {\bf{\hat Y}}} \right\|_2^2 + {C_L}{\left\| {{\bf{W}}_{{\mathop{\rm out}\nolimits} }^{}} \right\|_1} + C\left\| {{\bf{W}}_{{\mathop{\rm out}\nolimits} }^{}} \right\|_2^2,
\end{equation}
the advantage of the proposed method lies in its output compensation mechanism. This scheme avoids direct output prediction and instead focuses on modelling the residuals, which not only enhances the model generalization performance but also preserves structural compactness. Through this design, the network can more effectively adapt to varying input patterns, thereby reducing the risk of overfitting.

Furthermore, researchers can explore more effective methods for order variable extraction and temporal data analysis. This may include developing new algorithms to more accurately identify key order variables, as well as the integration of reservoir computing and deep learning techniques to handle complex time series data. Such efforts will enhance the performance and applicability of the model, thus increasing its competitiveness in practical applications.

\section{Conclusion}
Time-varying input orders and data distributions are prevalent in nonlinear dynamic systems within industrial processes. This paper presents an enhanced RSCN with hybrid regularization for problem solving. The LASSO-RSCN-L2 model integrates the strengths of LASSO and L2 regularization, enabling it to effectively extract relevant order variable information while preventing overfitting in the presence of unknown data. The universal approximation property and echo state property of the network are also ensured during the construction. Experimental results indicate that the proposed method can significantly alleviate ill-posedness and demonstrates considerable potential in nonlinear dynamic modelling.

Notably, the LASSO-RSCN-L2 framework lacks a theoretical basis for selecting the regularization coefficient $C$. Future research could focus on developing a data-dependent $C$ to enhance the adaptability and performance of the model, especially in scenarios with varying noise levels and complexity. Moreover, subsequent studies may extend this strategy to block incremental RSCNs \cite{ref30}, further improving the learning efficiency and effectiveness of the network.

\end{CJK}
\end{document}